\documentclass{article}

\usepackage{microtype}
\usepackage{graphicx}
\usepackage{booktabs} 

\usepackage{hyperref}

\usepackage[accepted]{icml2023}

\usepackage{mathtools}

\usepackage[capitalize,noabbrev]{cleveref}

\usepackage{amsmath,amsfonts,bm}

\def\eqref#1{equation~\ref{#1}}

\def\1{\bm{1}}

\DeclareMathAlphabet{\mathsfit}{\encodingdefault}{\sfdefault}{m}{sl}
\SetMathAlphabet{\mathsfit}{bold}{\encodingdefault}{\sfdefault}{bx}{n}

\newcommand{\sigmoid}{\sigma}

\usepackage{hyperref}
\usepackage{url}
\usepackage{tabularx}
\usepackage{booktabs}
\usepackage{multirow}
\usepackage{hhline}
\usepackage{amsmath, amsthm, amssymb}
\usepackage{booktabs}
\usepackage{amssymb}
\usepackage{amsmath}
\usepackage{array, makecell}
\usepackage{graphicx}
\usepackage{mathrsfs}  
\usepackage{enumitem}
\usepackage{tikz}
\usepackage[T1]{fontenc} 
\usepackage[ruled,vlined,algo2e]{algorithm2e}
\usepackage{algpseudocode}
\usepackage{algorithm}
\usepackage{wrapfig}
\usepackage{soul,xcolor}
\usepackage{caption}
\usepackage{subcaption}
\usepackage{stackengine}
\usepackage{diagbox}
\setstcolor{magenta}

\makeatletter
\newcommand{\omgset}[3][0ex]{%
  \mathrel{\mathop{#3}\limits^{
    \vbox to#1{\kern-2\ex@
    \hbox{$\scriptscriptstyle#2$}\vss}}}}
\makeatother

\makeatletter
\newcommand{\uset}[3][0ex]{%
  \mathrel{\mathop{#3}\limits_{
    \vbox to#1{\kern-6\ex@
    \hbox{$\scriptscriptstyle#2$}\vss}}}}
\makeatother

\makeatletter
\newcommand{\oset}[2]{%
  {\mathop{#2}\limits^{\vbox to 1.5\ex@{\kern-\tw@\ex@
  \hbox{\scriptsize #1}\vss}}}}
\makeatother

\newcommand{\Pmc}{\mathcal{P}}
\newcommand{\Xmc}{\mathcal{X}}
\newcommand{\Rmcstar}{\mathcal{R}^*}
\newcommand{\Pto}{{\omgset{*}{\uset{\Pmc}{\Rightarrow}}}}

\newcommand{\metagrammar}{\overline{G}}
\newcommand{\ted}{\mathit{TED}}

\newcommand{\PGmetato}{{\omgset{*}{\uset{{\Pmc}_{\overline{G}}}{\Rightarrow}}}}
\newcommand{\pto}{{{\oset{$p$}{\Rightarrow}}}}
\newcommand{\pito}{{{\oset{$p_i$}{\Rightarrow}}}}

\newcommand\myeq{\mkern3.0mu{=}\mkern3.0mu}

\theoremstyle{plain}
\newtheorem{theorem}{Theorem}[section]
\newtheorem{proposition}[theorem]{Proposition}

\theoremstyle{definition}
\newtheorem{definition}[theorem]{Definition}

\theoremstyle{remark}

\usepackage[textsize=tiny]{todonotes}

\icmltitlerunning{Hierarchical Grammar-Induced Geometry for Data-Efficient Molecular Property Prediction}

\begin{document}

\twocolumn[
\icmltitle{Hierarchical Grammar-Induced Geometry for \\ Data-Efficient Molecular Property Prediction}



\icmlsetsymbol{equal}{*}

\begin{icmlauthorlist}
\icmlauthor{Minghao Guo}{MIT CSAIL}
\icmlauthor{Veronika Thost}{IBM}
\icmlauthor{Samuel W Song}{MIT CSAIL}
\icmlauthor{Adithya Balachandran}{MIT CSAIL}
\\
\icmlauthor{Payel Das}{IBM}
\icmlauthor{Jie Chen}{IBM}
\icmlauthor{Wojciech Matusik}{MIT CSAIL}
\end{icmlauthorlist}

\icmlaffiliation{MIT CSAIL}{MIT CSAIL}
\icmlaffiliation{IBM}{MIT-IBM Watson AI Lab, IBM Research}

\icmlcorrespondingauthor{Minghao Guo}{guomh2014@gmail.com}

\icmlkeywords{Machine Learning, ICML}

\vskip 0.3in
]



\printAffiliationsAndNotice{}  

\begin{abstract}
The prediction of molecular properties is a crucial task in the field of material and drug discovery.  
The potential benefits of using deep learning techniques are reflected in the wealth of recent literature. Still, these techniques are faced with a common challenge in practice: Labeled data are limited by the cost of manual extraction from literature and laborious experimentation. 
In this work, we propose a data-efficient property predictor by utilizing a learnable hierarchical molecular grammar that can generate molecules from grammar production rules. 
Such a grammar induces 
an explicit geometry of the space 
of molecular graphs, which provides an informative prior on molecular structural similarity. 
The property prediction is performed using graph neural diffusion over the grammar-induced geometry.
On both small and large datasets, our evaluation shows that this approach outperforms a wide spectrum of baselines, including 
supervised and pre-trained graph neural networks.
We include a detailed ablation study and further analysis of our solution, showing its effectiveness in cases with extremely limited data.
Code is available at \url{https://github.com/gmh14/Geo-DEG}.
\end{abstract}

\section{Introduction}
Molecular property prediction is an essential step in the discovery of novel materials and drugs, as it applies to both high-throughput screening and molecule optimization.
Recent advances in machine learning, particularly deep learning, have made tremendous progress in predicting property values that are difficult to measure in reality due to the associated cost. 
Depending on the representation form of molecules, various methods have been proposed, including recurrent neural networks (RNN) for SMILES strings~\citep{lusci2013deep, goh2017smiles2vec}, feed-forward networks (FFN) for molecule fingerprints~\citep{tao2021benchmarking, tao2021machine}, and, more dominantly, graph neural networks (GNN)
for molecule graphs~\citep{yang2019analyzing, bevilacqua2021equivariant, aldeghi2022graph, yu2022molecular}. They have been employed to predict biological and mechanical properties of both polymers and drug-like molecules. 
Typically, these methods learn a deep neural network that maps the molecular input into an embedding space, where molecules are represented as latent features and then transformed into property values.
Despite their promising performance on common benchmarks, these deep learning-based approaches require a large amount of training data in order to be effective~\citep{audus2017polymer, wieder2020compact}.

In practice, however, scientists often have 
small datasets at their disposal, in which case deep learning fails.
Many recent studies have demonstrated that handling small dataset scenarios is a non-trivial open research problem and is of practical importance which has consistently attracted the attention of the fields of molecule and material discovery~\citep{subramanian2016computational, altae2017low, audus2017polymer}.
For example, due to the difficulty of generating and acquiring data -- which usually entails synthesis, wet-lab measurement, and mechanical testing -- state-of-the-art works on polymer property prediction using real data are limited to only a few hundred samples~\citep{menon2019hierarchical, chen2021polymer}.
Recent deep learning research
has developed various methods handling small molecular datasets, 
including self-supervised learning~\citep{zhang2021motif, rong2020self, wang2022molecular, ross2021large}, transfer learning~\citep{hu2019strategies}, and few-shot learning~\citep{guo2021few, stanley2021fs}. 
These methods involve pre-training networks on large molecular datasets before being applied to domain-specific, small-scale target datasets. 
However, when applied to datasets of very small size (e.g., ${\sim}300$), most of these methods 
are prone to perform poorly and are statistically unstable ~\citep{hu2019strategies}. 
Moreover, as we will show in our experiments, these methods are less reliable when deployed on target datasets that contain significant domain gaps from the pre-trained dataset (e.g., inconsistency in molecule sizes).

As an alternative to pure 
deep learning-based methods, formal grammars over molecular structures offer an explicit, explainable representation for molecules and have shown their great potential in addressing molecular tasks in a data-efficient manner~\citep{kajino2019molecular, krenn2019selfies, guo2021polygrammar, guo2022data}.
A molecular grammar consists of a set of production rules that can be chained to generate molecules.
The production rules, which can either be manually defined~\citep{krenn2019selfies, guo2021polygrammar} or learned from data~\citep{kajino2019molecular, guo2022data}, encode necessary constraints for generating valid molecular structures, such as valency restrictions.
A molecular grammar has the combinatorial capacity to represent a large amount of molecules using a relatively small number of production rules. It has thus been adapted as a data-efficient generative model~\citep{kajino2019molecular, guo2022data}.
While molecular generation based on grammars has been widely studied, extending the data-efficiency advantage of grammars to property prediction poses a new challenge that has not yet been well-explored: 
Grammar-based property prediction, particularly for small datasets, requires learning how to map molecule space to potentially divergent property spaces, a task that is significantly more challenging than exploring within the molecule space as the generative model does. 

\begin{figure*}[tb]
\centerline{\includegraphics[width=0.9\linewidth]{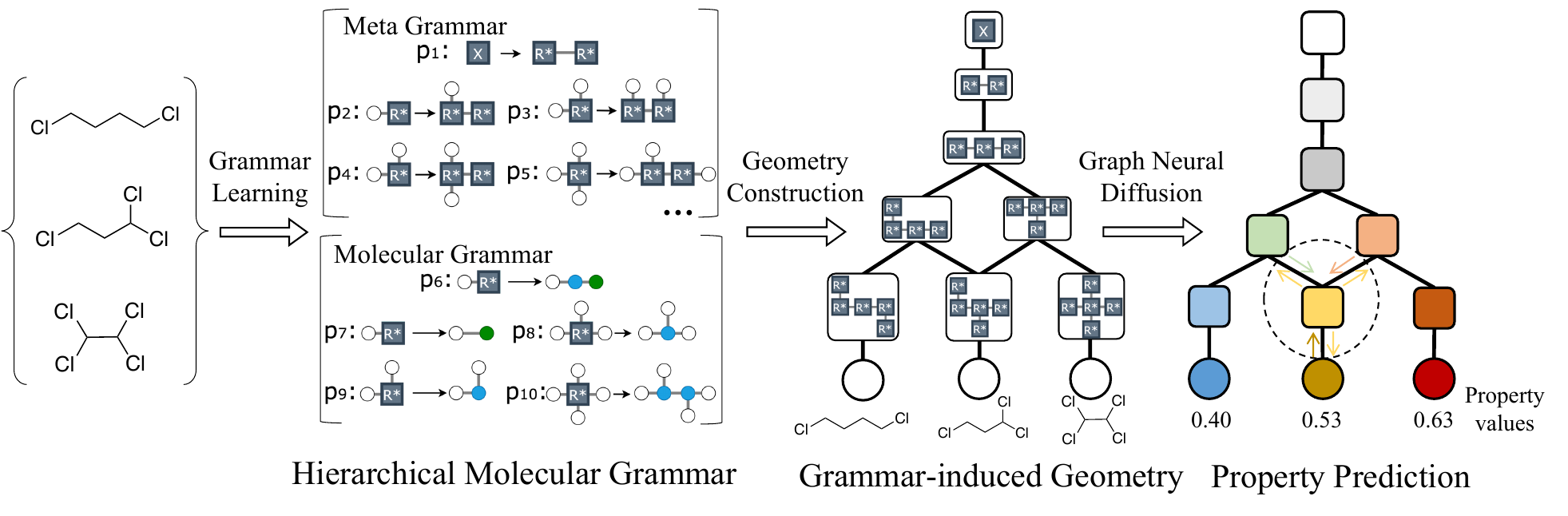}}
\vspace{-2.5ex}
\caption{Overview. Given a set of molecules, we learn a hierarchical molecular grammar that can generate molecules from production rules. 
The hierarchical molecular grammar induces an explicit geometry for the space of molecules, where structurally similar molecules are closer in distance along the geometry. Such a grammar-induced geometry provides an informative prior for data-efficient property prediction.
We achieve this by using graph neural diffusion over the geometry.}
\vspace{-1.ex}
\label{fig:motivation}
\end{figure*}

\textbf{Motivation.} 
In this paper, we propose a framework
for highly data-efficient property prediction based on a learnable molecular grammar.
We approach this problem not by trivially deploying a separate property predictor model on top of the grammar, but instead, investigating the underlying reason why most deep-learning methods fail in data-sparse cases and identifying the key motivation for integrating the grammar into our framework: A grammar can provide an \emph{explicit prior} on the similarity of molecular structures.
For instance, two molecular structures that share a common substructure would use the same sequence of grammar production rules.
As it is widely recognized in cheminformatics that molecules with similar structures have similar properties~\citep{johnson1990concepts, martin2002structurally}, grammar production sequences can thus be used as a strong structural prior to predict molecular properties.  
We aim to develop a model that explicitly represents grammar production sequences and captures the structure-level similarity between molecules.
Even from few molecules, this model is expected to reveal a wealth of information regarding the structural relationship between molecules, providing the key to a data efficient property predictor.


\textbf{Framework.} 
Figure~\ref{fig:motivation} outlines our approach.
At the heart of our method is a grammar-induced \emph{geometry} (in the form of a graph) for the space of molecules.
In the geometry, every path tracing from the root to a leaf
represents a grammar production sequence that generates a particular molecule.
Such a geometry explicitly captures the intrinsic closeness between molecules, where structurally similar molecules are closer in distance along the geometry. 
In contrast to the embedding space used in most deep learning methods, our grammar-induced geometry is entirely \emph{explicit}, which can be integrated with the property predictor by considering all molecules in the space simultaneously.
However, it is infeasible to construct this geometry using existing grammars, since they incur a prohibitively high computational cost as we later demonstrate in the analysis.
We propose a \emph{hierarchical molecular grammar} consisting of two parts: a pre-defined \emph{meta grammar} at the top and a learnable \emph{molecular grammar} at the bottom.
Our hierarchical molecular grammar, without any loss of expressiveness, is the first grammar that can be used to construct a geometry at runtime. 
We provide both solid theoretical and experimental evidence to demonstrate that the hierarchical molecular grammar is compact yet complete.
To predict properties, we exploit the structural prior captured by the grammar-induced geometry using a graph neural diffusion model over the geometry.
A joint optimization framework is proposed to learn both the geometry and the diffusion in an end-to-end manner.

\textbf{Evaluation.}
Our evaluation covers $8$ commonly used benchmark datasets, including classification and regression problems, with datasets of both small (${\sim}300$) and large ($600{-}1,000$) sizes.
Our approach significantly outperforms both competitive state-of-the-art GNN approaches and fine-tuned pretrained models.
Further analysis shows that when trained on only a subset of the training data (${<}100$), our method can achieve performance comparable to pre-trained GNNs fine-tuned on the whole training set of the downstream prediction task, thus demonstrating the effectiveness of our method on extremely small datasets. 

\textbf{Contributions.} \textbf{1)} Our framework, to the best of our knowledge, is the first method that leverages the data efficiency of a learnable molecular grammar for the task of molecular property prediction through learning the geometry of the space of molecules. \textbf{2)} Our hierarchical molecular grammar takes the inherent advantages of general molecular grammars, including explicitness, explanatory power, and data efficiency, but expands the utility and is more compact than all existing grammars. \textbf{3)} We show that our method achieves significantly better performance on challenging small datasets and outperforms a wide spectrum of baselines on various common benchmarks.




\textbf{Related Works.}
Our method is mainly related to three areas: 1) machine learning, particularly graph neural networks for molecular property prediction, in which we target the same problem but propose an entirely new approach; 
2) grammars for molecular machine learning, which are used in existing works for molecular generation, while we use grammars to induce a geometry for molecular property prediction;
and 3) geometric deep learning, where deep networks are applied to non-Euclidean domains. 
We briefly discuss the latter two in Section~\ref{sec:preliminaries} as necessary background for our approach and refer to Appendix~\ref{app:related} for an in-depth discussion.

\textbf{Differences Compared to~\citet{guo2022data}.}
Our work offers key improvements over~\citet{guo2022data} in multiple aspects. 
First, we develop a solution for property prediction, a task different from molecular generation and one that lacks a straightforward solution as to how grammars are applied. 
Second, we extend non-trivially a grammar to a hierarchical grammar, with crucial attributes -- $k$-degree, edit-complete, and minimal -- that provide greater expressiveness and compactness compared to the grammar in~\citet{guo2022data}. 
Third, our hierarchical grammar is generic and independent of specific molecule datasets. This allows us to construct the (meta-)geometry of the molecular graphs once in a lifetime, which is infeasible for existing, non-hierarchical grammars. 
Overall, our approach surpasses~\citet{guo2022data} in both theory and practice, achieving compelling performance in property prediction tasks compared with a wide variety of baselines. 

\section{Preliminaries}
\label{sec:preliminaries}

\textbf{Molecular Hypergraph Grammar (MHG).} 
In MHGs, molecules are represented as hypergraphs $H\myeq(V,E)$, where nodes represent chemical atoms and hyperedges represent chemical bonds or ring-like substructures.
A MHG $G\myeq(\mathcal{N}, \Sigma, \mathcal{P}, \mathcal{X})$ contains a set $\mathcal{N}$ of non-terminal nodes, a set $\Sigma$ of terminal nodes representing chemical atoms, and a starting node $\mathcal{X}$. 
It describes how molecular hypergraphs are generated 
using a set of production rules $\mathcal{P}\myeq\{p_i|i=1,...,k\}$ of form $p_i: \mathit{LHS} \rightarrow \mathit{RHS}$, where the left- and right-hand sides ($\mathit{LHS}$ and $\mathit{RHS}$) are hypergraphs.
Starting at $\mathcal{X}$, a molecule is generated by iteratively selecting a rule whose $\mathit{LHS}$ matches a sub-hypergraph in the current hypergraph and replacing it with $\mathit{RHS}$ until only terminal nodes remain.
For each production rule, the $\mathit{LHS}$ contains only non-terminal nodes 
and no terminal nodes, whereas the $\mathit{RHS}$ can contain both non-terminal and terminal nodes. 
Both sides of the production rule have the same number of anchor nodes, which indicate correspondences when $\mathit{LHS}$ is replaced by $\mathit{RHS}$ in a production step.
For a formal definition, see~\cite{guo2022data}.

\textbf{Graph Diffusion} models the information propagation between nodes on a graph using heat equations.
%
The node features are updated following a diffusion PDE as follows:
\begin{align}\label{eq:diffusion}
\begin{split}
    & \mathbf{U}_T = \mathbf{U}_0 + \int_0^T\frac{\partial \mathbf{U}_t}{\partial t}\text{d}t, \quad \quad
   \frac{\partial \mathbf{U}_t}{\partial t} = \mathbf{L}_\alpha\mathbf{U}_t,
\end{split}
\end{align}
where matrix $\mathbf{U}_t$ represents the features of all nodes in the graph at time $t$ and the matrix $\mathbf{L}_\alpha$, which has the same sparsity structure as a graph Laplacian,
represents the diffusivity defined on all edges in the graph.
$\mathbf{L}_\alpha$ is calculated using $a(\cdot, \cdot; \alpha)$ parameterized by $\alpha$, i.e. $L_{ij} = a(U_t^{\smash{(i)}}, U_t^{\smash{(j)}}; \alpha)$ for all connected node pairs $(i,j)$.
For more details, see \cite{chamberlain2021grand}. 
%


\textbf{Notation.} 
For a hypergraph 
grammar $G=(\mathcal{N}, \Sigma, \mathcal{P}, \mathcal{X})$, we say a graph $H$ can be \emph{derived} from the grammar if there is a sequence of production rules from $\mathcal{P}$ that generates this graph, denoted by $\Xmc\Pto H$.
$H_1\pto H_2$ denotes one-step grammar production that transforms graph $H_1$ to graph $H_2$ using rule~$p$.
As a special form of general graph, a tree is denoted as $T$ and the set of all possible trees is denoted as $\mathcal{T}$.
Note that all trees discussed in this paper are \emph{unrooted without order}, i.e. connected acyclic undirected graphs~\citep{bender2010lists}. 
$\Delta(T)$ denotes the maximal degree of $T$.
$\mathit{TED}(T_1, T_2)$ denotes the \emph{tree edit distance} between tree $T_1$ and tree $T_2$ as defined in~\cite{zhang1996constrained} and \cite{paassen2018revisiting}.
$|\mathcal{P}|$ denotes the number of rules in a production rule set $\mathcal{P}$. $|T|$ denotes the size of a tree $T$.


\section{Grammar-induced Geometry for Molecular Property Prediction}
\label{sec:approach}

\subsection{General Grammar-induced Geometry}

\textbf{Problem Formulation.} 
A molecular property predictor can be expressed as a function $\pi(\cdot): \mathcal{H} \mkern3.0mu{\rightarrow}\mkern3.0mu \mathbb{R} $ that maps molecules formulated as hypergraphs $H\myeq(V, E) \in \mathcal{H}$ into scalar values.
The mapping function $\pi = g \circ f$ contains two components: an embedding function $f(\cdot): \mathcal{H} \mkern3.0mu{\rightarrow}\mkern3.0mu\mathbb{R}^n$ that maps input hypergraphs into a Euclidean \emph{latent feature space} (also known as the \emph{embedding space}), and a transformation function $g(\cdot): \mathbb{R}^n \mkern3.0mu{\rightarrow}\mkern3.0mu \mathbb{R}$ (usually a simple linear function) which maps latent features into property values.
The embedding function is designed to capture the similarity between molecules, such that molecules with similar properties are closer in the embedding space in terms of Euclidean distance~\citep{cayton2005algorithms}.
The key reason why most supervised machine learning methods fail in data-sparse cases is that the embedded function fails to capture molecular similarity since it only has access to a limited number of samples with property labels.
To address this issue, we propose an additional, explicit prior on molecular similarity that does not depend on data labels, but provides valuable information for property prediction.

\begin{figure}[tb]
\centerline{\includegraphics[width=1.0\linewidth]{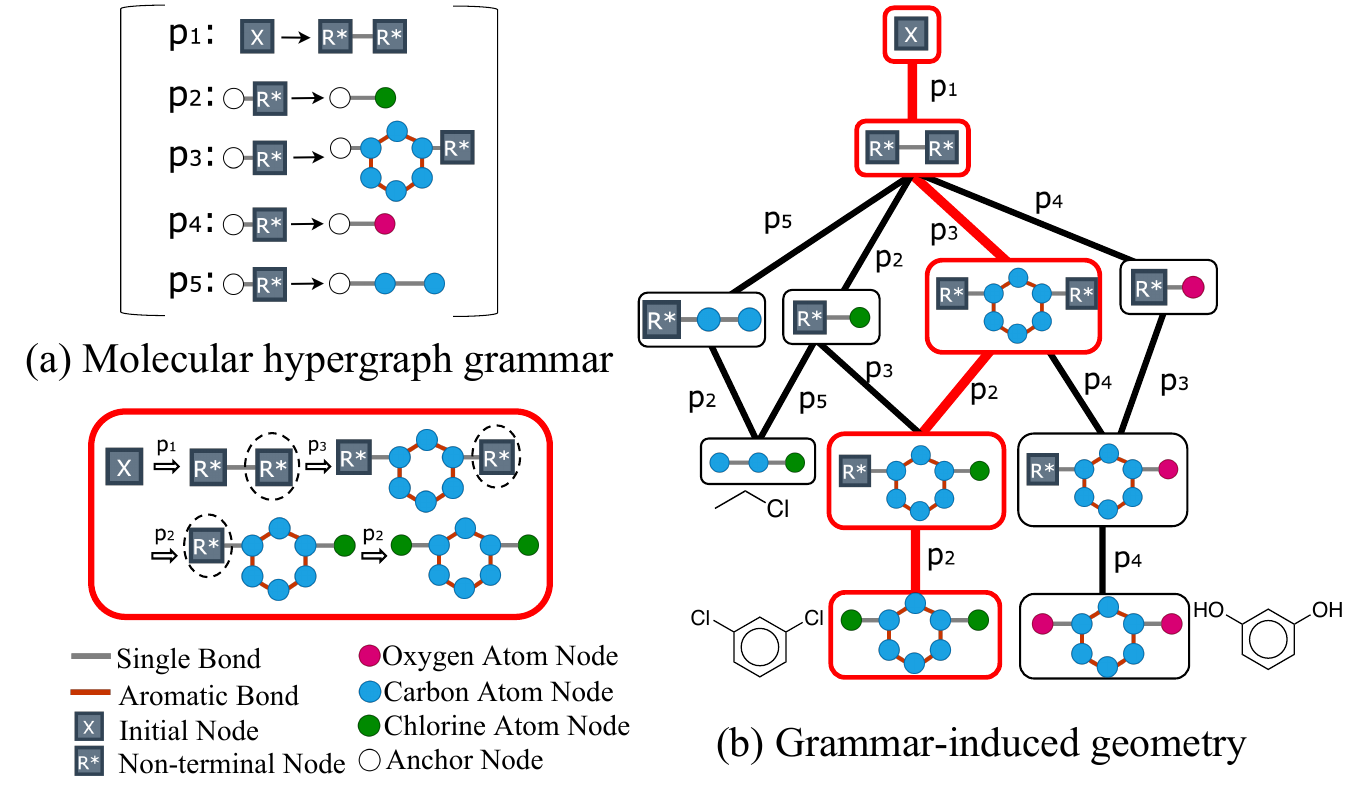}}
\vspace{-1ex}
\caption{(a) Example of a molecular hypergraph grammar and (b) its induced geometry. All possible production sequences to generate three molecules are outlined in the geometry, where the nodes represent intermediate or molecular hypergraphs and the edges represent production steps. 
Structurally similar molecules have a smaller shortest path between them and are closer in distance along the geometry.}
\label{fig:pipeline}
\vspace{-3.5ex}
\end{figure}

\textbf{General Grammar-induced Geometry.} \label{sec:geometry}
Our efforts are motivated by the fact that 
\emph{structure-level similarity} between molecules is known to be strongly correlated with their properties, e.g., organic molecules sharing a functional group are proven to have similar chemical properties ~\citep{johnson1990concepts, martin2002structurally}. 
We \emph{explicitly} model this structure-level similarity as an informative prior for data-efficient property prediction.

As grammars are generative models for molecules constructed by combining molecular substructures, we exploit the molecular hypergraph grammar to capture structural similarity between molecules.
Figure~\ref{fig:pipeline} outlines how a molecular hypergraph grammar $G\myeq(\mathcal{N}, \Sigma, \mathcal{P}, \mathcal{X})$ can be used to construct a \emph{geometry} for the space of molecules in the form of a graph $\mathcal{G}\myeq(\mathcal{V}, \mathcal{E})$\footnote{To distinguish the graph of the geometry from the graph of molecules, we use \emph{hypergraphs} to formulate the molecular graphs generated from grammar production.}.
The geometry enumerates all possible production sequences, where every possible intermediate or molecular hypergraph is represented as a node in $\mathcal{V}\myeq \{v|v\myeq H_v(V_v, E_v)\}$, and each production step is represented as an edge in $\mathcal{E}\myeq\{(s,t)|H_s\vphantom{H}\pto H_t, p\in\mathcal{P}\}$.
In this geometry, every leaf node represents a molecule. 
Every path tracing from the root $H_{root} \myeq (\mathcal{X}, \varnothing)$ to a leaf represents a production sequence that generates the corresponding molecule.
Since all the molecules can be derived from the initial node $\mathcal{X}$, any two leaves representing two molecules are connected by at least one common ancestor (i.e. the root) in this geometry, where the path between them represents the sequence of steps that would be required to transform from one to the other.
Molecules with greater structural similarity have a smaller \emph{shortest path} as their molecular hypergraphs share a common intermediate hypergraph.
The distance between two molecules is defined as the shortest-path distance between them along the geometry with unit weight at each edge.
We use this geometry as an additional input to the molecular property predictor: $\pi{'}(\cdot, \cdot): \mathcal{H} \times \mathfrak{G} \rightarrow \mathbb{R}$, where $\mathcal{G}\in\mathfrak{G}$ is the grammar-induced geometry.
The geometry can be optimized in conjunction with the property predictor in order to minimize the prediction loss. 
As the geometry is determined by the grammar, the optimization of the geometry can be converted into the learning of production rules, where the latter can be achieved using the method described in \cite{guo2022data}.

The crucial remaining question is how to construct this geometry $\mathcal{G}\myeq(\mathcal{V}, \mathcal{E})$ from a given molecular hypergraph grammar.
A key characteristic of the grammar-induced geometry is that each node in the geometry represents a \emph{unique} intermediate hypergraph or molecular hypergraph.
This ensures that the geometry includes the minimal path between two molecules.
To satisfy this characteristic, one trivial idea of geometry construction is to use breadth-first search (BFS). 
Originating from the root $H_{root} \myeq (\mathcal{X}, \varnothing)$, the geometry can be iteratively expanded following the production rules until all the molecules involved in the property prediction task have been visited.
In practice, however, such a method of constructing grammar-induced geometry is very costly, and often computationally intractable for more complex grammars.
By testing with random grammars learned from \cite{guo2022data}, we find that it is infeasible to construct the geometry when there are more than ten production rules.
A detailed analysis is provided in Appendix~\ref{app:constructcost}.
The major bottleneck comes from the combinatorial complexity of production rules: The more rules a grammar has, the more intermediate hypergraphs it can generate.
As the depth of the geometry increases, the number of intermediate hypergraphs increases exponentially.
This incurs a significant computational cost, as each newly expanded hypergraph requires pair-wise isomorphism tests.
This cost also poses a serious obstacle to the optimization of the geometry, since the geometry must be rebuilt at each optimization iteration.

\subsection{Hierarchical Molecular Grammars}\label{sec:hiergrammardag}
We propose a \emph{hierarchical molecular grammar} to address the computational challenge mentioned above.
Our insight stems from the fact that every hypergraph can be represented as a tree-structure object, i.e. a \emph{junction tree} using tree decomposition.
A junction tree is constructed by contracting certain vertices of the hypergraph into a single node so that it is cycle-free~\citep{diestel2005graph}.
The generation process of a molecular hypergraph can thus be divided into two parts: first generating a tree with homogeneous tree nodes, then converting the tree into a molecular hypergraph by specifying a sub-hypergraph per tree node.
As circle-free graphs, trees are more compact, so the possible tree structures are considerably fewer than molecular hypergraphs, and therefore can be enumerated.
The key to efficient grammar-induced geometry construction is that 
the enumeration of all possible homogeneous tree structures is data-independent and can be computed offline.
The only computational cost of constructing the geometry at run-time is attributed to converting trees into molecular hypergraphs.

\begin{figure*}[tb] 
\centerline{\includegraphics[width=0.95\linewidth]{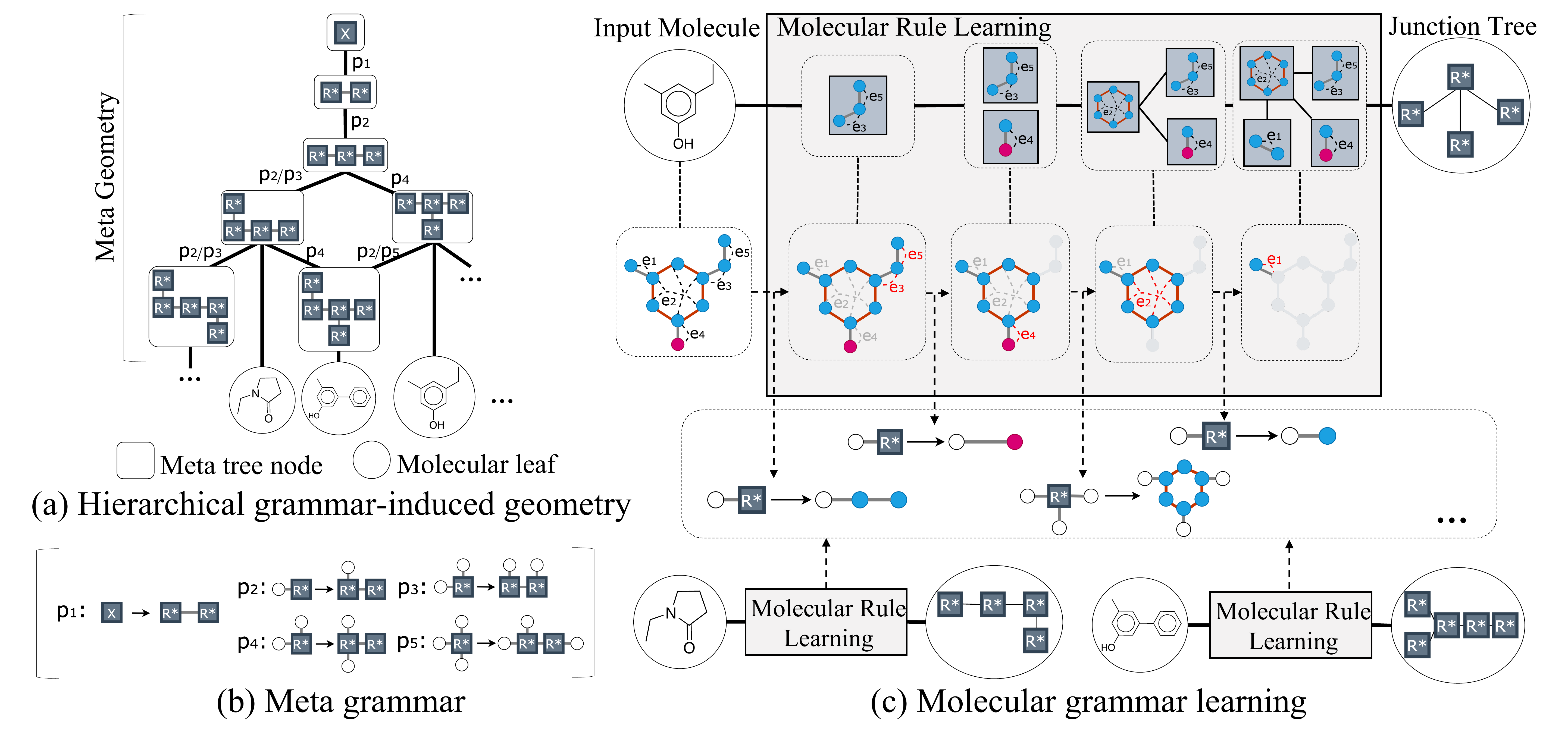}}
\vspace{-2.5ex}
\caption{
(a) Overview of the geometry induced by hierarchical molecular grammar. 
(b) A pre-defined meta grammar is used to pre-compute the meta geometry offline.
(c) At run-time, molecular grammar is obtained using molecular rule learning. 
Each molecule is converted into a junction tree and is connected to a meta tree node of the meta geometry that represents an isomorphic meta tree. 
}
\label{fig:metagrammarJT}
\vspace{-1.5ex}
\end{figure*}

We design a hierarchical molecular grammar to fulfill this principle.
A hierarchical molecular grammar consists of two sub-grammars: a \emph{meta grammar} that only generates trees, and a \emph{molecular grammar} that can convert trees into molecular hypergraphs.
The grammar-induced geometry can thus be divided into two parts: The top part (referred to as \emph{meta geometry}) is constructed by the meta grammar, where geometry nodes represent only trees, while the bottom part (referred to as \emph{molecular leaves}) is constructed by the molecular grammar, where geometry leaves represent molecular hypergraphs.
Figure~\ref{fig:metagrammarJT} shows an overview of a hierarchical molecular grammar and its induced geometry.
Thanks to this hierarchical decomposition, the meta geometry can be pre-computed to apply to any molecular datasets.
At run-time, we only need to construct the bottom part of the geometry, 
which is formulated as a byproduct of grammar learning in our approach.
As a result of the geometry construction, each molecule is connected to one junction tree in the meta geometry. 
Molecular structure similarity is determined by the distance between their corresponding junction trees along the geometry.
We provide a formal definition of meta grammar and demonstrate that despite the additional hierarchical constraint, our hierarchical molecular grammar is \emph{as expressive as} the general molecular hypergraph grammar in \cite{guo2022data}. 

\textbf{Meta Grammars.} The definition of meta grammar is,
\begin{definition}\label{def:metagrammar}
A \emph{meta grammar} $\metagrammar=(\mathcal{N}, \varnothing, \Pmc_{\metagrammar}, \mathcal{X})$ is a hypergraph grammar, which only contains non-terminal nodes 
and only generates trees, i.e. $\forall w\in \{\Xmc\PGmetato w\}, w\in\mathcal{T}$. \\
A meta grammar $\metagrammar$ is \emph{$k$-degree} if, for all trees $T$ of maximal degree $\Delta(T) \leq k$, 
we have
  $\Xmc\PGmetato T$.\\
A meta grammar~$\metagrammar$ 
is \emph{edit-complete} if, for any tree pair $(T, T')$ with $|T|<|T'|$, tree edit distance $\ted(T,T')=1$, and $\Xmc\PGmetato T,T'$, 
there is a rule $p\in\Pmc_{\metagrammar}$ such that $T\pto T'$.\\ 
%
A $k$-degree, edit-complete meta grammar~$\metagrammar$ is \emph{minimal} if there is no other such 
meta grammar $\metagrammar'$ with~$|\Pmc_{\metagrammar'}|<|\Pmc_{\metagrammar}|$. 
%
\end{definition}%
In Appendix~\ref{app:metarules}, we provide the formal construction of meta grammars for generating trees with arbitrary degree and elaborate on the three additional attributes.
\begin{proposition}
A meta grammar $\metagrammar$ with a meta rule set $\Pmc_{\metagrammar}$ as constructed in Appendix~\ref{app:metarules} is $k$-degree (for some $k$), edit-complete, and minimal. 
\end{proposition}
Appendix~\ref{app:metarules} also provides the 
proof that the three attributes are satisfied.
Generally speaking, we construct meta rule sets of arbitrary degrees by induction from the $1$-degree meta grammar, which consists of only one meta rule. 
In practice, a meta grammar with a reasonable degree should be chosen to ensure the coverage of most molecules.
In our experiments, we find it sufficient to use a meta rule set of degree $4$, which contains $8$ rules in total.

\textbf{Hierarchical Molecular Grammars.} 
We define the hierarchical molecular grammar as follows,
\begin{proposition}\label{prop:hmg}
A \emph{hierarchical molecular grammar} $\mathbf{G}=(\metagrammar, G_{mol})$---consisting of two hypergraph grammars: 1) a $k$-degree, edit-complete, and minimal meta grammar $\metagrammar$ defined in Definition~\ref{def:metagrammar}; and
2) a \emph{molecular grammar} $G_{mol}=(\mathcal{N}, \Sigma, \Pmc_{mol}, \varnothing)$ where the $\mathit{RHS}$ of any molecular rule in $\Pmc_{mol}$ only contains terminal nodes (i.e. atoms)
---is a \emph{complete} grammar.
\end{proposition}
More discussion on Proposition~\ref{prop:hmg} is in Appendix~\ref{app:metarules}. The completeness of our hierarchical molecular grammar also shows that there is no loss of representation capacity as compared to general molecular hypergraph grammars.

\subsection{{Hierarchical Grammar-induced Geometry}}

\textbf{Geometry Construction.}\label{sec:geometryconstruct}
The geometry induced by the hierarchical molecular grammar is constructed in two phases: Figure~\ref{fig:metagrammarJT}(a) shows the geometry constructed by the meta grammar at the top, and the geometry constructed by the molecular grammar at the bottom.
Since the meta grammar is pre-defined, the top meta geometry can be precomputed following the depth-limited BFS procedure described in Sec.~\ref{sec:geometry}.
Each node in the meta geometry represents a tree of non-terminal nodes.
We call the tree generated using the meta grammar a \emph{meta tree} and a node in the meta geometry a \emph{meta tree node}.
We find it sufficient to use a maximum BFS depth of $10$ in practice.
The bottom part of the geometry determines how each molecule is connected to the meta geometry as a {molecular leaf} through its junction tree structure.
We obtain the junction trees of molecules as a result of the learning of molecular grammar rules.
Thus, the geometry optimization can be achieved by learning the molecular grammar.

\textbf{Molecular Grammar Learning.}\label{sec:grammarlearning}
Figure~\ref{fig:metagrammarJT}(c) illustrates the process of molecular grammar learning.
We follow the grammar learning from~\cite{guo2022data} but constrain the learned rules to contain only terminal nodes on the $\mathit{RHS}$,  so as to satisfy the molecular grammar.
The molecular rules are constructed along with the junction tree.
Specifically, each molecule is considered as a hypergraph.
At each iteration, a set of hyperedges is sampled from the hypergraph.
The iterative sampling follows an i.i.d. Bernoulli distribution based on a probability function $\phi(e;\theta)$ defined on each hyperedge $e$ with learnable parameters $\theta$.
For each connected component of these sampled hyperedges, we construct a molecular rule, where the $\mathit{LHS}$ contains a single non-terminal node and the $\mathit{RHS}$ contains the connected component.
For the junction tree,
we add a junction tree node representing the connected component and create a junction tree edge between two junction tree nodes if their corresponding connected components share a common hypergraph node.
Then the sampled hyperedges are removed from the original hypergraph.
The process terminates until all hyperedges are removed. 
At the conclusion of the sampling process, we can obtain a junction tree of the molecular hypergraph along with a set of molecular rules.
The molecule is then connected to the meta tree node in the meta geometry that represents the isomorphic meta tree to the junction tree. 
We provide a detailed formulation of the constructed molecular rule and the resulting junction tree in Appendix~\ref{app:molrules} and more details of the geometry construction in Appendix~\ref{app:grammardag}.

\subsection{Molecular Property Prediction}
\textbf{Graph Diffusion on Grammar-induced Geometry.}\label{sec:graphdiff}
Property prediction requires a model that can operate on our grammar-induced geometry and is suitable for scarce data.
We choose the graph neural diffusion model GRAND~\citep{chamberlain2021grand} for its effectiveness in overcoming the oversmoothing that plagues most GNNs.
Three learnable components are used in a graph diffusion process: an encoder function~$\varphi$ defined on all the nodes in the grammar-induced geometry, a decoder function~$\psi$ defined on molecular leaves, and a graph diffusion process given in Eq.~\ref{eq:diffusion}. 
Specifically, the encoder~$\varphi$ yields the initial state of the diffusion process $\mathbf{U}_0 = \varphi(\mathbf{U}_\text{in})$, where $\mathbf{U}_\text{in}$ is the matrix form of the input features of all the nodes. 
The decoder produces predicted property values of all molecular leaves $\mathbf{u}_T = \psi(M\odot\mathbf{U}_T)$, where $\mathbf{U}_T$ is the node-feature matrix from the final diffusion state, $M$ is a binary mask that masks out the rows corresponding to non-molecule nodes, $\odot$ is the Hadamard product, and $\mathbf{u}_T$ is the resulting vector containing property values of all molecular leaves.
Our optimization objective includes the learning of both the grammar-induced geometry and the diffusion model:
\begin{align}\label{eq:opt}
\begin{split}
&\min_{\theta, (\varphi, \psi, \alpha)}{l}(\mathbf{u}_T,  \hat{\mathbf{u}}) = \min_{\theta}\min_{(\varphi, \psi, \alpha)}{l}(\mathbf{u}_T,  \hat{\mathbf{u}}),
\end{split}
\end{align}
where $\hat{\mathbf{u}}$ represents the vector of ground-truth property values for all leaves and $l(\cdot, \cdot)$ is a regression loss. 
Recall that the geometry is only determined by the molecular rules, so the molecular grammar learning parameters $\theta$ are the only parameters relevant for obtaining the geometry. 
Since $\theta$ and $(\varphi, \psi, \alpha)$ are two groups of independent variables, we exploit block coordinate descent to optimize the objective.
A detailed derivation is provided in Appendix~\ref{app:optimization}.

\textbf{Overall Framework of End-to-end Learning.}
First, we compute the meta geometry offline using the predefined meta grammar (Sec.~\ref{sec:hiergrammardag}). Then, given a set of molecules, we use molecular grammar learning (Sec.~\ref{sec:grammarlearning}) to convert each molecule into a junction tree and attach it to the meta geometry to form the whole hierarchical grammar-induced geometry (Sec.~\ref{sec:geometryconstruct}). Graph diffusion (Sec.~\ref{sec:graphdiff}) is then applied to the geometry to obtain a scalar value for each molecular leaf, i.e. predicted property value for each molecule. Training is conducted by minimizing the error between the predicted and ground-truth values (Eq.~\ref{eq:opt}). The parameters of molecular grammar learning and graph diffusion are trained in an end-to-end manner.
A graphical illustration of the optimization framework is provided in Appendix~\ref{app:optimization}.

\textbf{Application to Transductive and Inductive Setting.}
Our system can be used in two different ways.
The first is to use our framework as both a property predictor and a generative model through the use of learned hierarchical molecular grammar, which is the default setting in our experiments.
To achieve this, our framework takes all molecules as input (from both training and testing datasets), but only uses the property values of training datasets, forming a \emph{transductive} setting.
This ensures that all the input molecules can be generated using the learned grammar.
The other setting is to use our framework as a pure property predictor.
In this case, we ignore the learned grammar rules but consider the molecular rule learning module as a function to convert \emph{any} input molecule (both in and out of the training dataset) into a junction tree, which is then attached to the meta geometry.
Thus, our framework does not require molecules from testing datasets as input, forming an \emph{inductive} setting.
Specifically, we use the same training procedure as for the transductive case, except we only use molecules from the training dataset as input.
When testing a new molecule, by using the trained grammar sampling function $\phi(e;\theta)$ and following the hyperedge contraction process, we can obtain the junction tree for that molecule. 
Upon attachment to the isomorphic meta tree node, the new molecule is added to grammar-induced geometry. The property value of the molecule can be predicted following the graph diffusion using the trained decoder~$\varphi$ and encoder~$\phi$.

\section{Evaluation}

The experiments demonstrate the \textbf{generality of our approach} and answer the following questions:
1) Does our approach outperform existing methods 
on \textbf{small datasets}?
2) How well does our approach perform on \textbf{large, widely-studied datasets}?
3) To what extent is our approach effective on \textbf{extremely small datasets}?
4) Is our approach effective for \textbf{both transuctive and inductive} settings?


\subsection{Experiment Setup}
\textbf{Data.}
{
We evaluate our approach on eight datasets: CROW (a curated dataset from literature), Permeability~\citep{yuan2021imputation}, FreeSolv~\citep{mobley2014freesolv}, Lipophilicity~\citep{wang2015silico}, HOPV~\citep{lopez2016harvard}, DILI~\citep{ma2020deep}, PTC~\citep{xu2018powerful}, and ClinTox~\citep{gayvert2016data}.
These datasets cover: 1)  commonly used benchmark datasets including MolecuNet (FreeSolv, Lipophilicity, HOPV, ClinTox) and TUDataset (PTC), 2) both classification (DILI, PTC, ClinTox) and regression tasks (CROW, Permeability, FreeSolv, Lipophilicity, HOPV), and 3) sizes that are small (CROW, Permeability, HOPV, PTC, DILI) and large (FreeSolv, Lipophilicity, ClinTox).
}
{
We report
mean absolute error (MAE) and coefficient of determination ($R^2$) 
for regression, and Accuracy and AUC for classification. 
See Appendix~\ref{app:datasets} for details.
}

\textbf{Baselines.}
We compare our approach with various approaches: Random Forest, FNN, wD-MPNN (D-MPNN), ESAN, HM-GNN, PN, and Pre-trained GIN. For descriptions, see Appendix \ref{app:implementation}.
To show the generality of our pipeline (called Geo-DEG), we implement two versions, each with a different diffusion encoder, GIN and MPNN.
Appendix~\ref{app:implementation} provides the implementation details.

\begin{table*}[t]
\centering
\scriptsize
\caption{Results on $6$ out of $8$ datasets (best result \textbf{bolded}, second-best \underline{underlined}).
More results on the other two datasets are in Appendix~\ref{app:moreresults}.
Our approach (Geo-DEG) outperforms both state-of-the-art supervised and semi-supervised GNNs.}
\vspace{-1.5ex}
\label{table:small}
\begin{tabular}{@{}>{\centering\arraybackslash}p{1cm}>{\centering\arraybackslash}p{1cm}|cc|ccc|cc|cc@{}}
\toprule
\multicolumn{2}{c|}{\multirow{2}{*}{\backslashbox{Datasets}{Methods}}} & \multirow{2}{*}{\shortstack{Random\\Forest}} & \multirow{2}{*}{FFN} & \multirow{2}{*}{wD-MPNN} & \multirow{2}{*}{ESAN} & \multirow{2}{*}{HM-GNN} & \multirow{2}{*}{\shortstack{PN\\(finetuned)}} & \multirow{2}{*}{\shortstack{Pre-trained GIN\\(finetuned)}} & \multirow{2}{*}{\shortstack{\textbf{Geo-DEG}\\ \textbf{(GIN)}}} & \multirow{2}{*}{\shortstack{\textbf{Geo-DEG}\\ \textbf{(MPNN)}}}\\
  & & & & & & & & & &\\
\midrule
\multirow{2}{*}{\textbf{CROW}} & \textbf{MAE}\ $\downarrow$ & 27.9 ± 3.2 & 24.0 ± 2.1 & 20.6 ± 1.3 & 26.1 ± 1.3 & 30.8 ± 1.8 & 21.1 ± 1.3 & 19.3 ± 1.4 & \textbf{17.0 ± 1.4} & \underline{18.5 ± 1.2} \\
&$\mathbf{R^2}\uparrow$ & 0.67 ± 0.08 & 0.84 ± 0.02 & 0.89 ± 0.02 & 0.79 ± 0.02 & 0.76 ± 0.01 & 0.89 ± 0.01 & \underline{0.91 ± 0.01} & \textbf{0.92 ± 0.01} & \underline{0.91 ± 0.01} \\
\midrule
\multirow{2}{*}{\textbf{Permeability}} & \textbf{MAE}\ $\downarrow$ & 0.58 ± 0.01 & 0.56 ± 0.04 & 0.46 ± 0.03 & 0.40 ± 0.03 & 0.49 ± 0.03 & 0.48 ± 0.04 & 0.46 ± 0.04 & \underline{0.34 ± 0.02} & \textbf{0.32 ± 0.03} \\
&$\mathbf{R^2}\uparrow$ & 0.72 ± 0.03 & 0.73 ± 0.06 & 0.80 ± 0.03 & 0.81 ± 0.03 & 0.68 ± 0.01 & 0.70 ± 0.03 & 0.69 ± 0.02 & \textbf{0.84 ± 0.02} & \underline{0.83 ± 0.02} \\
\midrule
\multirow{2}{*}{\textbf{FreeSolv}} & \textbf{MAE}\ $\downarrow$ & 4.58 ± 0.34 & 3.67 ± 0.40 & \underline{0.54 ± 0.08} & 0.67 ± 0.07 & 0.63 ± 0.05 & 0.65 ± 0.04 & 1.01 ± 0.11 & 0.62 ± 0.06 & \textbf{0.49 ± 0.06} \\
&$\mathbf{R^2}\uparrow$ & 0.60 ± 0.09 & 0.77 ± 0.05 & \underline{0.90 ± 0.02} & 0.86 ± 0.03 & 0.86 ± 0.02 & 0.85 ± 0.02 & 0.74 ± 0.03 & \underline{0.90 ± 0.02} & \textbf{0.94 ± 0.02} \\
\midrule
\multirow{2}{*}{\textbf{Lipophilicity}} & \textbf{MAE}\ $\downarrow$ & 0.64 ± 0.04 & 0.51 ± 0.02 & \underline{0.44 ± 0.02} & 0.46 ± 0.04 & 0.58 ± 0.04 & 0.56 ± 0.04 & 0.52 ± 0.03 & 0.48 ± 0.02 & \textbf{0.42 ± 0.02} \\
&$\mathbf{R^2}\uparrow$ & 0.65 ± 0.03 & 0.80 ± 0.03 & \underline{0.90 ± 0.02}  & \underline{0.90 ± 0.01} & 0.84 ± 0.02 & 0.81 ± 0.02 & 0.89 ± 0.01 & 0.88 ± 0.02 & \textbf{0.91 ± 0.02} \\
\midrule
\multirow{2}{*}{\textbf{HOPV}} & \textbf{MAE}\ $\downarrow$ & 0.36 ± 0.03 & 0.35 ± 0.03 & 0.36 ± 0.03 & 0.37 ± 0.02 & 0.40 ± 0.02 & 0.42 ± 0.02 & 0.38 ± 0.02 & \underline{0.32 ± 0.03} & \textbf{0.30 ± 0.02} \\
&$\mathbf{R^2}\uparrow$ & 0.69 ± 0.05 & 0.67 ± 0.06 & 0.69 ± 0.04 & 0.66 ± 0.06 & 0.65 ± 0.05 & 0.65 ± 0.04 & 0.66 ± 0.03 & \underline{0.70 ± 0.03} & \textbf{0.74 ± 0.03} \\
\midrule
\multirow{2}{*}{\textbf{PTC}} & {\textbf{Acc.}\ $\uparrow$} & 0.60 ± 0.06 & 0.58 ± 0.06 & 0.67 ± 0.06 & 0.64 ± 0.08 & \underline{0.66 ± 0.07} & 0.61 ± 0.08 & 0.62 ± 0.09 & 0.64 ± 0.09 & \textbf{0.69 ± 0.07} \\
&{\textbf{AUC}$\uparrow$} & 0.63 ± 0.05 & 0.61 ± 0.04 & 0.70 ± 0.05 & 0.68 ± 0.06 & \underline{0.69 ± 0.06} & 0.65 ± 0.07 & 0.66 ± 0.07 & 0.68 ± 0.06 & \textbf{0.71 ± 0.07} \\
\bottomrule
\end{tabular}%
\vspace{-2.5ex}
\end{table*}

\subsection{Results on Small Datasets}

\textbf{Results \& Discussion.}
To address question 1), we conduct experiments on five small datasets: CROW, Permeability, HOPV, PTC, and DILI.
Table \ref{table:small} shows the results on the first four datasets and Appendix~\ref{app:moreresults} shows the results on DILI.
Both variants of our method outperform all the other methods by a large margin.
For the two polymer datasets, CROW is a more challenging dataset than Permeability, as it has fewer samples and a broader range of property values (with a standard derivation of $86.8$ versus $1.24$ for Permeability).
On CROW, traditional machine learning methods such as Random Forest and FNN are quite competitive and even outperform modern GNN architectures such as ESAN and HM-GNN.
wD-MPNN achieves reasonable performance thanks to the special graph representation for polymer ensembles.
PN and Pre-trained GIN perform exceptionally well on CROW due to their pre-training on large datasets.
However, this pre-training does not help on Permeability, which has a much larger average molecule size (with an average molecular weight of $391.8$ versus $153.8$ for CROW). 
Thus, there is a domain gap between the dataset to pre-train these networks and the Permeability dataset used for testing, resulting in poor performance of both methods.
ESAN benefits from the subgraph selection scheme on large molecular graphs and performs well on Permeability.
The poor performance of HM-GNN on both datasets shows that it is not as effective in regression as it is in classification.
The overall results show that: 1) our method can handle molecules with varying sizes, and 2) the superior performance of our method when coupled with either GIN or MPNN confirms its generalizability to different feature extractors.
More discussion on the other datasets are given in Appendix~\ref{app:moreresults}.

\subsection{Results on Large Datasets } 

\textbf{Results \& Discussion.}
To address question 2), we conduct experiments on three large datasets: FreeSolv, Lipophilicity, and ClinTox.
Table~\ref{table:small} and \ref{table:clintox} show the results. 
Our method equipped with MPNN diffusion encoder performs the best among all approaches.
On large datasets, random forest and simple FFN become less competitive than on small datasets.
This is reasonable since larger datasets require larger model capacities to represent the relationship between molecules and their properties.
Pre-training does not significantly help on large datasets due to the inherent domain gap between datasets used for pre-training versus those used for fine-tuning.
Among the three GNN-based baselines, D-MPNN performs the best and exhibits better stability than ESAN and HM-GNN.
Our approach can further improve the performance of D-MPNN and outperforms all the baselines.
Results and discussion on the other datasets are given in Appendix~\ref{app:moreresults}.
These results demonstrate that our method is scalable for large molecule datasets.

\subsection{Analysis}

\begin{figure}[tb]
\centerline{\includegraphics[width=0.9\linewidth]{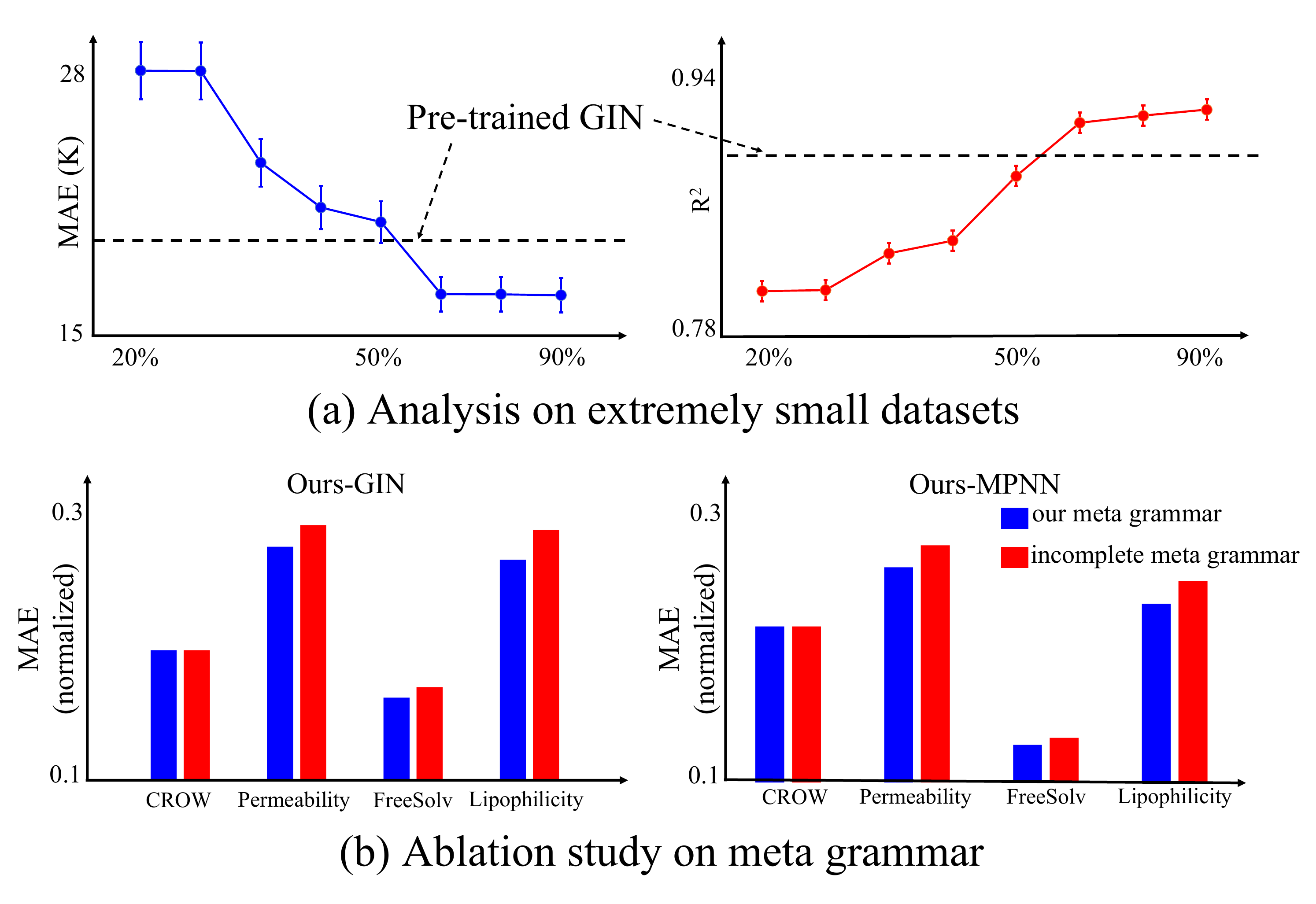}}
\vspace{-2ex}
\caption{
{(a) Performance of our method trained on subsets of CROW training data with different ratios. Even when the training set is halved, our approach can still achieve competitive results compared to Pre-trained GIN \emph{fine-tuned on the whole training set}.}
(b) Comparison of performance using different meta grammars, demonstrating the necessity of a complete meta grammar. }
\label{fig:exp}
\vspace{-3ex}
\end{figure}

\textbf{Performance on Extremely Small Datasets.} \label{sec:analysis}
We conduct a study on the minimum viable number of samples for our method.
We randomly sample $20\%$ of CROW as a fixed testing set.
Using the remaining data, we construct eight randomly selected training sets, each consisting of a portion of all remaining data, ranging from $20\%$ to $100\%$.
These eight training sets are used to train our method using GIN as the feature extractor.
Figure~\ref{fig:exp}(a) shows the performance on the testing set.
Even when the training set is halved (with only $94$ samples), our approach still achieves results that are comparable to those of the Pre-trained GIN fine-tuned on the whole training set. Appendix~\ref{app:sizedep} includes a more detailed study 
for the effect of changing dataset training size on model performance.

\textbf{Performance on Both Transductive and Inductive Settings.}
We compare both the transductive and inductive settings of our method on the CROW dataset (the most challenging scarce dataset).  
Figure~\ref{table:transductive} in the Appendix shows the results.
Our method achieves similar results when applied inductively and transductively, both of which outperform the best semi-supervised inductive baseline--the finetuned Pretrained GIN--on CROW. 
Our results show that the performance gain of our method does not depend on the transductive setting, but rather on the learnable prior introduced by the grammar-induced geometry. 

\textbf{Ablation Study on Meta Grammar.}
In this study, we examine the necessity of the meta rules in our hierarchical grammar.
We remove two rules that have degree $4$ on the $\mathit{LHS}$ from the $4$-degree meta grammar, resulting in a meta geometry with same number of nodes but $10\%$ fewer edges than the one used in our main experiments.
With this modified meta grammar, we run the pipeline for four datasets and compare with the original meta grammar in Figure~\ref{fig:exp}(b).
All four datasets exhibit a performance drop when using the modified meta grammar.
The results provide experimental evidence for the necessity of a complete meta grammar.

\section{Conclusion}
We propose a data-efficient molecular property predictor based on a hierarchical molecular grammar. 
The grammar induces an explicit geometry describing the space of molecular graphs, such that a graph neural diffusion on the geometry can be used to effectively predict property values of molecules on small training datasets. 
One avenue of future work is to extend our pipeline to model 3D molecular structures and to address general graph design problems.

\section*{Acknowledgement}
This work is supported by the MIT-IBM Watson AI Lab, and its member company, Evonik.

\bibliography{example_paper}
\bibliographystyle{icml2023}

\newpage
\appendix
\onecolumn
\section{Additional Results} \label{app:moreresults}
\begin{table}[h]
\begin{minipage}{.5\textwidth}
  \centering
    \caption{Results on DILI.}
    \begin{tabular}{@{}ccc}
    \toprule
    \multirow{2}{*}{} 
    & \multicolumn{2}{c}{{\textbf{DILI}}} \\
                       &{\textbf{Accuracy}\ $\uparrow$}&{\textbf{ROC-AUC}$\uparrow$}\\ \midrule
    Random Forest      & 0.70 ± 0.09 & 0.80 ± 0.06 \\
    FFN                & 0.69 ± 0.07 & 0.78 ± 0.07 \\ \midrule 
    D-MPNN             & 0.75 ± 0.10 & 0.83 ± 0.07 \\
    ESAN               & 0.74 ± 0.10 & 0.82 ± 0.10 \\
    HM-GNN             & \underline{0.76 ± 0.08} & 0.83 ± 0.09 \\
    \midrule
    PN (finetued)                 & 0.75 ± 0.10 & 0.83 ± 0.06 \\
    Pre-trained GIN (finetued)   & 0.74 ± 0.09 & 0.82 ± 0.06 \\\midrule
    \textbf{Geo-DEG (GIN)}  & \underline{0.76 ± 0.09}  &  \underline{0.84 ± 0.05}  \\
    \textbf{Geo-DEG (MPNN)} & \textbf{0.78 ± 0.08}  &  \textbf{0.86 ± 0.06}   \\
    \bottomrule
    \end{tabular}%
\end{minipage}
\begin{minipage}{.5\textwidth}
  \centering
    \caption{Results on ClinTox.}
    \label{table:clintox}
    \begin{tabular}{@{}cc}
    \toprule
    {\textbf{Methods}}
                    &{\textbf{ROC-AUC}\ $\uparrow$}\\\midrule
    D-MPNN             &     90.6 ± 0.6   \\
    AttentiveFP        &     84.7 ± 0.3\\
    N-Gram$_{\text{RF}}$  &  77.5 ± 4.0\\
    N-Gram$_{\text{XGB}}$ &  87.5 ± 2.7\\
    GROVER$_{\text{base}}$&  81.2 ± 3.0\\
    GROVER$_{\text{large}}$& 76.2 ± 3.7\\
    GraphMVP          &      79.1 ± 2.8\\ 
    MolCLR            &      91.2 ± 3.5\\ 
    GEM               &      90.1 ± 1.3\\
    Pre-trained GIN    &     72.6 ± 1.5\\
    Uni-Mol           &      \underline{91.9 ± 1.8}\\ \midrule
    \textbf{Geo-DEG (GIN)}  & 74.4 ± 1.8  \\
    \textbf{Geo-DEG (MPNN)} & \textbf{92.2 ± 0.7}  \\
    \bottomrule
    \end{tabular}%
\end{minipage}
\vspace{-1ex}
\end{table}

\begin{figure}[bh]
  \begin{minipage}{0.50\linewidth}
    \begin{displaymath} 
      \begin{tabular}{l c c}
        \hline
        {{Metrics/Methods}} & \textbf{MAE}\ $\downarrow$ & $\mathbf{R^2}\uparrow$\\
        \hline
        \multirow{2}{*}{\shortstack{Pre-trained GIN \\ (fine-tuned)}} & \multirow{2}{*}{19.3 ± 1.4} & \multirow{2}{*}{0.91 ± 0.01} \\
        \\
        \multirow{2}{*}{\shortstack{Geo-DEG GIN \\ (transductive)}} & \multirow{2}{*}{17.0 ± 1.4} & \multirow{2}{*}{0.92 ± 0.01}\\
        \\
        \multirow{2}{*}{\shortstack{Geo-DEG GIN \\ (inductive)}}  & \multirow{2}{*}{17.2 ± 1.3} & \multirow{2}{*}{0.92 ± 0.01}\\
        \\
        \hline
    \end{tabular}
    \end{displaymath}
  \end{minipage}
  \begin{minipage}{0.50\linewidth}
    \centerline{\includegraphics[width=0.8\textwidth]{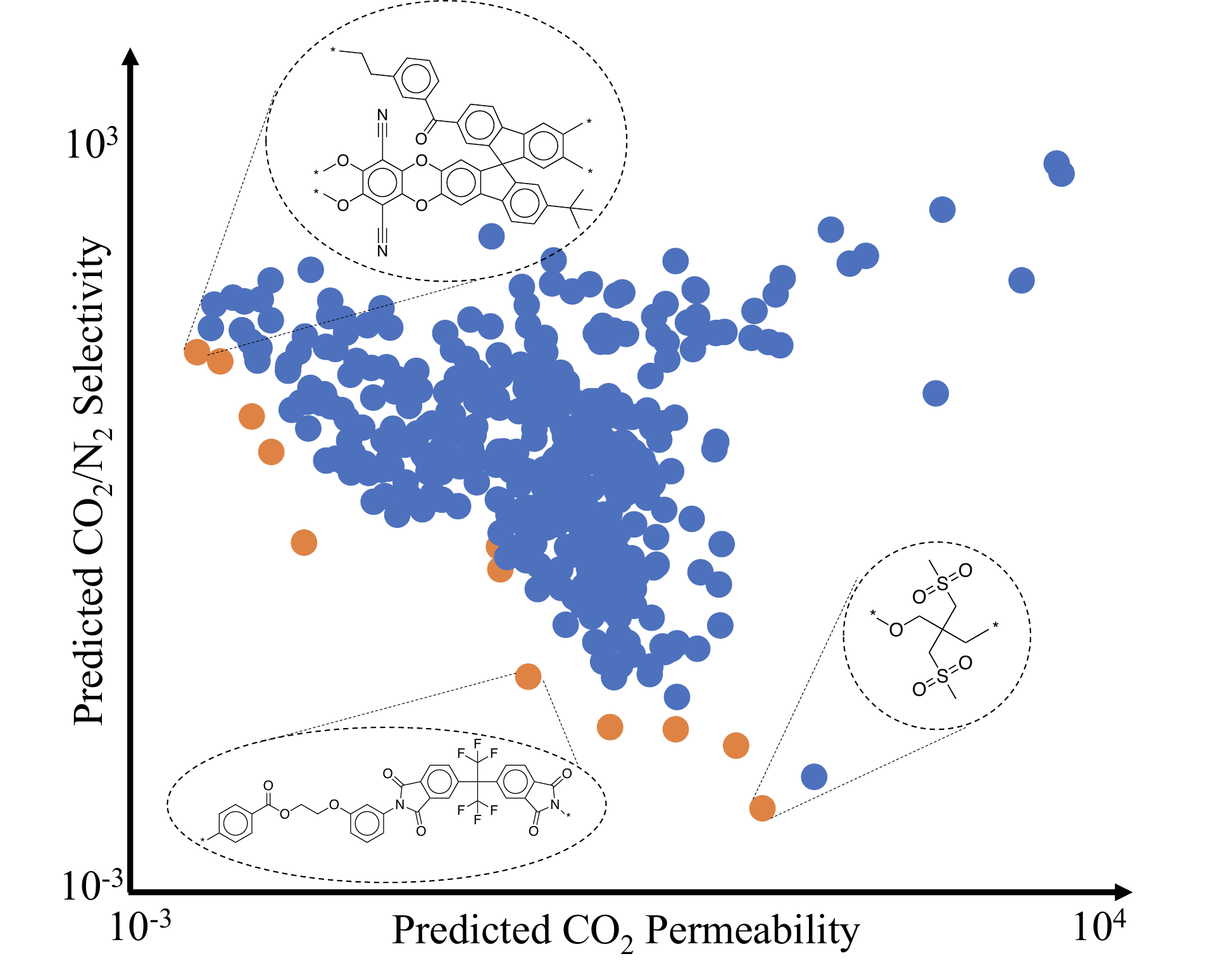}}
  \end{minipage} \hfill
  \caption{Comparison of transductive and inductive settings (left) and analysis on combining with generative models (right). Two objectives from the Permeability dataset are predicted using our method combined with the grammar-based generative model. Molecules lying on the Pareto front are highlighted.} \label{fig:alignment_exp}
  \label{table:transductive}
\end{figure}

\textbf{Results \& Discussion.}
For HOPV, since it is a small regression dataset, traditional methods (random forest and FFN) achieve competitive performance and are even better than several graph neural networks including ESAN and HM-GNN. Both variants of our method outperform other baselines by a large margin, which aligns with the results in the other two small regression datasets. 
In the two small classification datasets, DILI and PTC, there is no significant difference between baseline models regarding performance. The reason for this is that the test dataset is small, and the accuracy of the model can be limited by a few hard examples that cannot be classified correctly. It is therefore more informative to use the ROC-AUC score. Among the baselines, HM-GNN performs the best due to its motif-based representation. Our method can still outperform all other methods regarding both accuracy and ROC-AUC score.
For ClinTox in Table~\ref{table:clintox}, we compare our approach with a wide range of highly performant baseline methods from a recent paper~\cite{zhou2022uni}, which by itself proposes one of the state-of-the-art methods on ClinTox.
Note that these are very strong baseline methods, including many methods pre-trained using 3D molecular data.
The overall results show that our method achieves significantly better performance on challenging small regression datasets and outperforms a wide spectrum of baselines on various common benchmarks.

\section{Combination with Generative Models.}\label{app:transduct}
{
In our optimization framework, grammar metrics can be considered as additional objectives, allowing us to jointly optimize both generative models and property predictors.
Following~\cite{guo2022data}, we use diversity and Retro$^*$ score as our grammar metrics and then perform joint optimization on the Permeability dataset. 
After training, we generate $400$ molecules and predict their property values using our approach, including all six property types from Permeability.
Figure~\ref{table:transductive}(b) shows two out of six predicted property values with the Pareto front highlighted.
Clearly, our approach can be combined with generative models to provide a comprehensive pipeline for the discovery of optimal molecules. 
The use of Retro$^*$ also enables finding synthesis paths of generated molecules as shown in Appendix~\ref{app:retro}.}


\section{{Analysis on Training Dataset Size}}\label{app:sizedep}
\begin{figure*}[h]
\centerline{\includegraphics[width=0.8\linewidth]{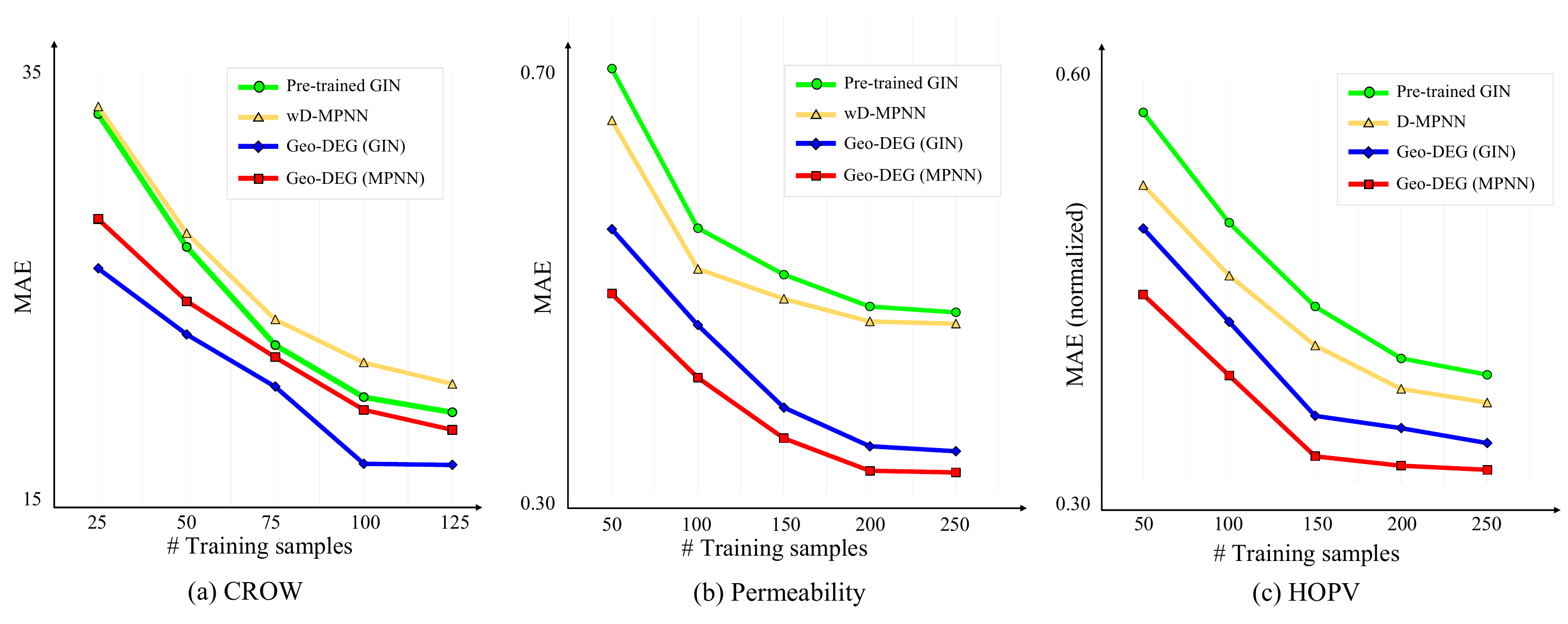}}
\vspace{-2ex}
\caption{{Analysis on training dataset size.}}
\label{fig:sizedep}
\vspace{-2ex}
\end{figure*}

{We further extend our study in Section~\ref{sec:analysis} for the effect of changing dataset training size on model performance by including the analysis of all three small regression datasets with more baselines compared. 
Figure~\ref{fig:sizedep} illustrates the results of Pre-trained GIN, wD-MPNN/D-MPNN, and two variants of our model on CROW, Permeability, and HOPV.
We report the performance of each model trained using different numbers of training samples, randomly sampled from the original training dataset.
Across different training dataset sizes, our proposed method consistently outperforms other baselines. A smaller training dataset leads to a larger performance improvement, demonstrating the data-efficiency of our model. 
}

\section{{Analysis on Construction Cost Reduction of Hierarchical Molecular Grammar}}\label{app:constructcost}
\begin{figure*}[t]
\centerline{\includegraphics[width=0.75\linewidth]{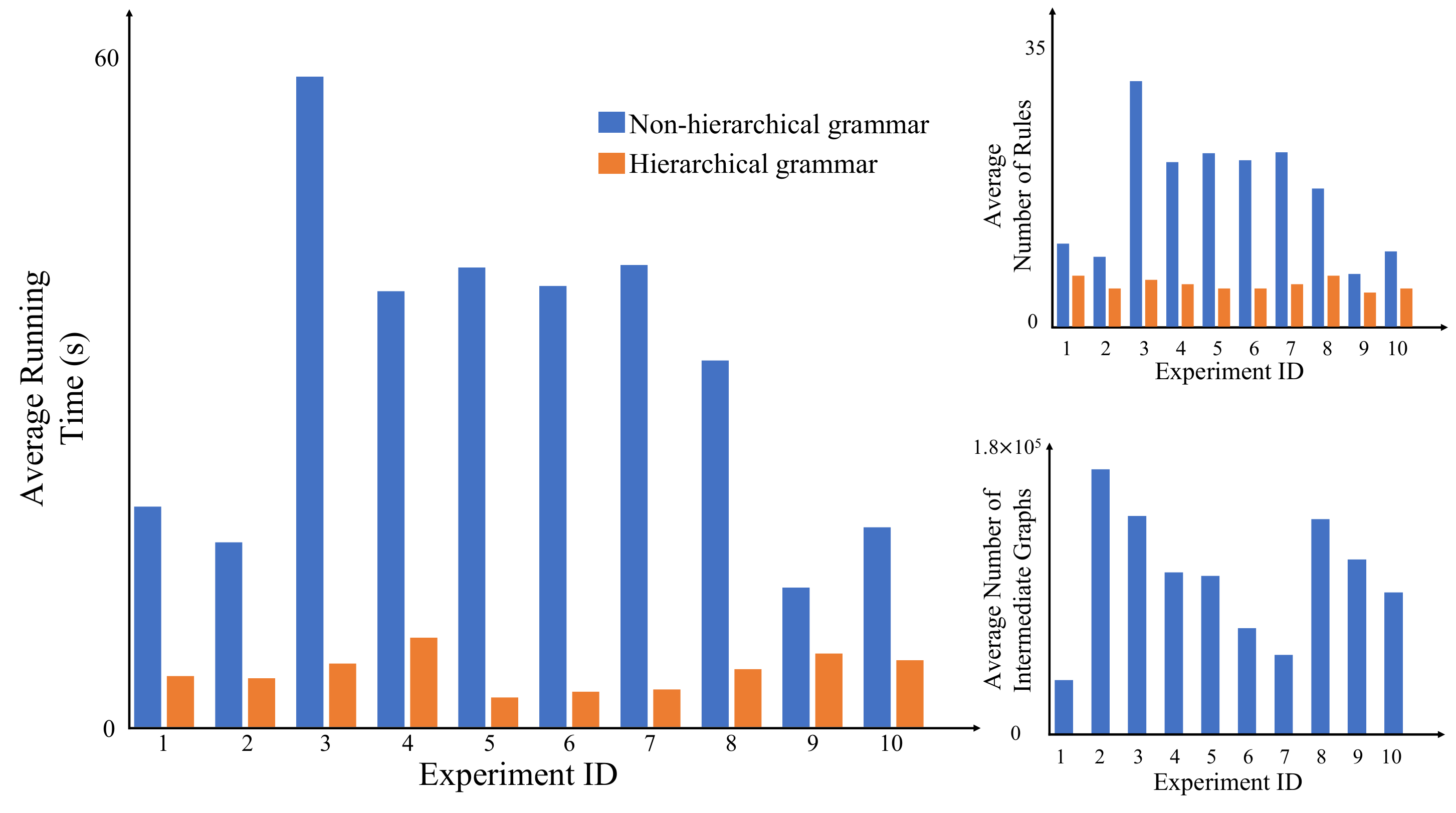}}
\vspace{-2ex}
\caption{{Analysis on construction cost reduction of hierarchical molecular grammar. For our hierarchical molecular grammar, the number of intermediate graphs is the same as the number of molecule samples, which is five in the experiments.}}
\label{fig:cost}
\vspace{-2ex}
\end{figure*}

{
In this section, we provide empirical evidence on the construction cost reduction of our hierarchical molecular grammar compared with non-hierarchical molecular grammar from \cite{guo2022data}.
We conduct ten groups of experiments by randomly sampling five molecules from CROW dataset by ten times.
For each group of experiment, we sample ten grammars for both hierarchical and non-hierarchical versions respectively and construct the geometry to cover the five molecules.
The non-hierarchical grammar is sampled using the algorithm from \cite{guo2022data}.
The geometry construction is achieved following the BFS procedure described in Section~\ref{sec:geometry} for both kinds of grammar.
In the experiments, we find that for non-hierarchical grammar, even for five molecules, it occurs frequently that the size of the geometry grows extremely large but still cannot cover the five molecules.
Therefore, we stop the construction when the size of the geometry reaches $2{\times}10^5$.
Figure~\ref{fig:cost} illustrates (a) the average running time, (b) the average number of production rules, and (c) the average number of intermediate graphs of the final constructed geometry.
It can be noted that our hierarchical molecular grammar greatly reduces the running time of the geometry construction.
For non-hierarchical grammars, it is always intractable to construct the geometry since the number of intermediate graphs is enormous (some approaching $1.5{\times}10^5$).
Our hierarchical molecular grammar provides a practical framework for grammar-induced geometry with theoretical foundations.
}

\section{{Analysis on Computational Complexity}}\label{app:compcomplex}
{The computation of our approach consists of three parts: 1) the construction of grammar-induced geometry, 2) diffusion encoder, and 3) graph diffusion over the geometry. 
Following \cite{blakely2021time}, suppose we have $K$ samples of molecules, each with $N$ nodes and $|\mathcal{E}|$ edges.
To construct the grammar-induced geometry, we need to sample all the edges in order to contract molecules into junction trees, so the computational complexity is $\mathcal{O}(K|\mathcal{E}|)$.
For diffusion encoder, the complexity is the same with general graph neural networks, which is $\mathcal{O}(KLd(Nd+ |\mathcal{E}|))$, where $L$ is the number of layeres and $d$ is the feature dimension.
As indicated in \cite{chamberlain2021grand}, the complexity of graph diffusion is $\mathcal{O}(|\mathcal{E}'|d)(E_b+E_f)$, where $\mathcal{E}'$ is the edge set of grammar-induced geometry, and $E_b$,  $E_f$ are the numbers of function evaluations for forward and backward pass, respectively.
Therefore, the overall computational cost of our method is $\mathcal{O}(K|\mathcal{E}| + KLd(Nd+ |\mathcal{E}|) +  |\mathcal{E}'|d(E_b+E_f)) = \mathcal{O}(KLd(Nd+ |\mathcal{E}|) + |\mathcal{E}'|d(E_b+E_f))$.
Compared with general graph neural networks which have a computational complexity of $\mathcal{O}(KLd(Nd+ |\mathcal{E}|))$, our approach entails additional computations $\mathcal{O}(|\mathcal{E}'|d(E_b+E_f))$, where $|\mathcal{E}'|$ is the sum of the number of edges in the meta geometry $|\mathcal{E}_{meta}|$ and the number of samples $K$, since each molecule is connected to the meta geometry by an edge.
In our experiments, we use a $4$-degree meta grammar with depth $10$, which gives us a meta geometry of $455$ edges, i.e. $455=|\mathcal{E}_{meta}|\ll K|\mathcal{E}|$. 
We also find in our experiments that $E_b + E_f \ll LNd$.
As a result, the overall computational complexity of our method is the same as that of general graph neural networks in the big $\mathcal{O}$ sense.
In practice, geometry can be constructed in parallel, which reduces the running time of our entire method.
}

\section{Details on Datasets}\label{app:datasets}
The Chemical Retrieval on the Web (CROW) polymer database\footnote{\url{https://polymerdatabase.com/}} curates a database of thermo-physical data for over $250$ different polymers, focusing primarily on the most commonly used plastics and resins in industry. CROW distinguishes from other digital polymer property databases due to its heavy usage of real, experimental data individually sourced via extensive literature search. In the CROW polymer database, approximately $95\%$ of reported polymers offer some experimental data from literature, with the remaining $5\%$ derived purely from simulation. 
We only use those data with labels from real experiments in our evaluation.
The CROW polymer database reports several properties including the glass transition temperature, Hildebrand solubility parameter, molar heat capacity, refractive index, and molar cohesive energy. Of these, we choose glass transition temperature as our benchmark performance comparison since it is one of the best documented polymer properties in the literature~\citep{tao2021benchmarking}.
For each dataset except ClinTox, we randomly split the data into $4:1$ training and testing sets and create five such splits using $5$ random seeds. 
For ClinTox, we follow the settings in \cite{zhou2022uni} to train our model and report the results of other baselines directly from the paper.
Each of five splits is used to train and test a separate instance of each benchmark model.
For the other three datasets, we refer the reader to the original papers: Permeability~\citep{yuan2021imputation}, FreeSolv~\citep{mobley2014freesolv}, Lipophilicity~\citep{wang2015silico}, HOPV~\citep{lopez2016harvard}, DILI~\citep{ma2020deep}, and PTC~\citep{xu2018powerful}.

\section{Details on the Implementation}\label{app:implementation}

\textbf{Baselines.} We compare our approach with various baselines: 1) Random Forest and FFN, two of the best-performing machine learning models for polymer informatics benchmarking~\citep{tao2021benchmarking}; 2) wD-MPNN, a state-of-the-art method specifically designed for polymer property prediction~\citep{aldeghi2022graph}; 3) ESAN, a general GNN architecture with enhanced expressive power~\citep{bevilacqua2021equivariant}; 4) HM-GNN, a motif-based GNN for molecular feature representation~\citep{yu2022molecular}; 5) PN and Pre-trained GIN, two pre-trained GNNs with state-of-the-art performance on few-shot learning~\citep{stanley2021fs} and transfer learning~\citep{hu2019strategies}. 
As HM-GNN and Pre-trained GIN only provide code for graph classification, we modify the final layer of their networks and use $l_1$ loss for training.
For the other methods, we follow the same implementation as their original papers.
Since wD-MPNN cannot be depolyed on general molecules other than polymers, we report the results of D-MPNN~\citep{yang2019analyzing} instead for two large datasets.

\textbf{Our System.}
For our approach, we use $4$-degree meta grammar, which contains eight rules.
The meta geometry contains all the meta trees whose size is smaller than $11$, resulting in $149$ nodes and $455$ edges.
For the molecular rule learning of $\theta$, we follow all the hyperparamters used in \cite{guo2022data}.
For the graph diffusion, the input feature of each meta tree node is the Weisfeiler Lehman graph hashing feature~\citep{shervashidze2011weisfeiler}.
The encoder for meta tree nodes is an embedding layer that maps hashing features into a $300$-dimension continuous vector.
For molecular leaves, we use two different encoders: GIN from \cite{xu2018powerful} and MPNN from \cite{yang2019analyzing}, both of which output a feature vector of dimension $300$.
For the decoder, we use a one-layer fully connected network with size $300$.
For the graph diffusion process, we follow \cite{chamberlain2021grand} and use Dormand–Prince adaptive step size scheme (DIORI5) with adjoint method.
The diffusivity function $a(\cdot, \cdot; \alpha)$ is an attention function.
We use Adam optimizer for the training of $\theta$ and $(\varphi, \psi, \alpha)$, with learning rate $0.01$ and $0.001$, respectively.
We train $\theta$ for ten epochs. 
For each training epoch of $\theta$, we train $(\varphi, \psi, \alpha)$ for $50$ epochs.

\section{Related works}\label{app:related}

\textbf{Machine Learning for Molecular Property Prediction.}
The use of machine learning methods to predict molecular properties has a long history.
Many early methods use SMILES strings or handcrafted fingerprints as input and rely on traditional machine learning methods, such as random forest and Gaussian processes, which are still competitive today in many applications~\citep{tao2021benchmarking, joo2022development}.
Recently, graph-based representations of molecules have gained increasing popularity with the development of GNNs~\citep{feinberg2018potentialnet, xu2018powerful, wu2021molformer, bevilacqua2021equivariant, yu2022molecular, aldeghi2022graph, alon2020bottleneck}. 
We refer the reader to~\cite{wieder2020compact} for a detailed review of GNN-based property predictors.
Current state-of-the-art GNN-based methods provide more advanced representations of molecules based on graphs. 
\cite{bevilacqua2021equivariant} represents each individual graph as a set of subgraphs, increasing the expressive power of GNNs.
\cite{aldeghi2022graph} tailors molecular graphs by adding stochastic edges and designs GNN specifically for polymers based on~\cite{yang2019analyzing}.
\cite{yu2022molecular} and \cite{zhang2021motif} leverage motifs to model molecular relationships. 
All these methods require large training datasets to achieve reasonable performance.
It is common to use self-supervised learning~\citep{rong2020self, wang2022molecular} and transfer learning~\citep{hu2019strategies} to deal with sparse data, where neural networks are pre-trained on large datasets and then fine-tuned on the target small dataset.
Most of these methods, however, focus solely on molecular classification, while our approach can also address regression problems with extremely sparse data, which are considerably more challenging. 

\textbf{Molecular Grammars.}
As an interpretable and compact design model, grammar has recently gained increasing attention in the field of molecule discovery~\citep{dai2018syntax, kajino2019molecular, krenn2019selfies, xu2020reinforced, nigam2021beyond, guo2021polygrammar, guo2022data}.
A molecular grammar can be considered as a generative model that uses production rules to generate molecules.
The rules of a grammar can be either manually constructed~\citep{dai2018syntax, krenn2019selfies, guo2021polygrammar} or automatically learned from molecular datasets~\citep{kajino2019molecular, xu2020reinforced, nigam2021beyond, guo2022data}.
Recent works have demonstrated that a learnable hypergraph grammar can be effective in capturing hard chemical constraints, such as valency restrictions~\citep{kajino2019molecular, guo2022data}.
\cite{guo2022data} further proposes a learning pipeline for constructing a hypergraph grammar-based generative model that can incorporate domain-specific knowledge from very small datasets with dozens of samples.
Despite the inherent advantages of molecular grammars, such as explicitness, explanatory power, and data efficiency, existing works mainly use them for molecular generation.
Prediction and optimization of molecular properties can only be accomplished by using a separate model based on the grammar representation in a latent space, which is learned individually~\citep{kajino2019molecular, xu2020reinforced}.
We integrate molecular grammar into property prediction tasks by constructing a geometry of molecular graphs based on the learnable grammar, which allows us to optimize both the generative model and the property predictor simultaneously.
On extremely small datasets, our approach benefits especially from the data efficiency of grammar and achieves superior performance over existing property predictors.

\textbf{Hierarchical Molecular Generation.}
The decomposition of molecular structures in our hierarchical molecular grammar is related to substructure-based methods in molecular generation~\citep{jin2018junction, jin2020hierarchical, maziarz2021learning}.
\cite{jin2018junction} employs encoders and decoders to generate a junction tree-structured scaffold and then combine chemical substructures into a molecule.
\cite{jin2020hierarchical} uses a hierarchical graph encoder-decoder to generate molecules in a coarse-to-fine manner, from atoms to connected motifs.
\cite{maziarz2021learning} integrates molecule fragments and atom-by-atom construction to generate new molecules.
Our approach is fundamentally different from all these existing hierarchical representations in two respects: 1) \emph{Without} the need for any training, our proposed meta grammar (the coarse level) can enumerate all possible junction tree structures by using a compact set of meta production rules, which can be \emph{theoretically guaranteed}.
Our method only requires learning at the fine level, which are molecular fragments determined by molecular rules, whereas existing methods require learning two models for both coarse and fine levels.
2) As opposed to existing methods that use latent spaces, our meta geometry induced by meta grammar explicitly models the similarity between molecules by using graph distance along the geometry.
Due to the edit-completeness of meta grammar in Definition \ref{def:metagrammar}, the graph distance is an explainable measurement of minimal editing distance between molecular graphs, whereas the distance in latent spaces used in existing methods lacks explainability.



\textbf{Geometric Deep Learning}
applies deep neural networks to non-Euclidean domains such as graphs and manifolds with a wide range of applications~\citep{bronstein2017geometric, cao2020comprehensive, bronstein2021geometric}.
Related to our method, a series of recent works use graph neural diffusion by treating GNNs as a discretisation of an underlying heat diffusion PDE~\citep{chamberlain2021grand, elhag2022graph, bodnar2022neural}.
These methods work on large graph data where the graph connectivity of individual nodes is provided.
Additionally, there are also existing works on inferring the underlying geometry of graph data.
\cite{ganea2018hyperbolic} and \cite{cruceru2021computationally} embed graphs into non-Euclidean manifolds such as hyperbolic and elliptical spaces and optimize their embeddings within these Riemannian spaces.
\cite{cosmo2020latent} learns a latent graph to model the underlying relationship between data samples and applies GNN to it.
Different from these methods, our approach uses explicit and learnable intrinsic geometry based on graph grammar to model the relationship between molecular data. 
Since graph grammar is a generative model, our framework is capable of optimizing both molecular generation and property prediction simultaneously. 

\clearpage
\section{More Details of Hierarchical Molecular Grammar}\label{app:metarules}

\textbf{Discussion on Hierarchical Molecular Grammar.}
The three additional attributes 
in Definition~\ref{def:metagrammar} can be used to deduce several desirable properties of a meta grammar.
``Degree $k$'' ensures the meta grammar is expressive and complete, covering all possible trees under a simple tree degree constraint.
``Edit completeness'' enables the explicit capture of transformations between two trees with edit distance one and therefore between two arbitrary trees with arbitrary distances.
``Minimality'' 
guarantees that the meta grammar is compact.
Based on these three attributes, we can construct a generic meta grammar generating trees of non-terminal nodes by using only a small but expressive set of production rules. 
Figure~\ref{fig:metagrammarJT}(b) shows a $3$-degree meta grammar. 
Each rule has one non-terminal node $\mathcal{R}^{*}$ on the $\mathit{LHS}$ and two $\mathcal{R}^{*}$s on the $\mathit{RHS}$.
Depending on the number of anchor nodes on the $\mathit{LHS}$, these rules can transform tree nodes of different degrees by attaching a new node using different schemes indicated by the $\mathit{RHS}$.
It is evident that we can generate any possible tree with a degree smaller than $4$ by adopting a sequence of rules from this $3$-degree meta grammar. 
In Proposition~\ref{prop:hmg}, ``Completeness'' states that any molecular graph can be derived from a hierarchical molecular grammar with a set of appropriate molecular rules.
This can be demonstrated by the fact that we can generate an arbitrary molecule by first using the meta grammar to generate a tree of non-terminal nodes (which has the same tree structure as a junction tree decomposed from the molecule), and then using the molecular grammar to transform the tree into a molecular hypergraph by specifying a molecular fragment for each non-terminal node. 

\begin{wrapfigure}{r}{0.40\textwidth}
\centering
\begin{minipage}{\linewidth}
    \centering\captionsetup[subfigure]{justification=centering}
    \includegraphics[width=0.9\linewidth]{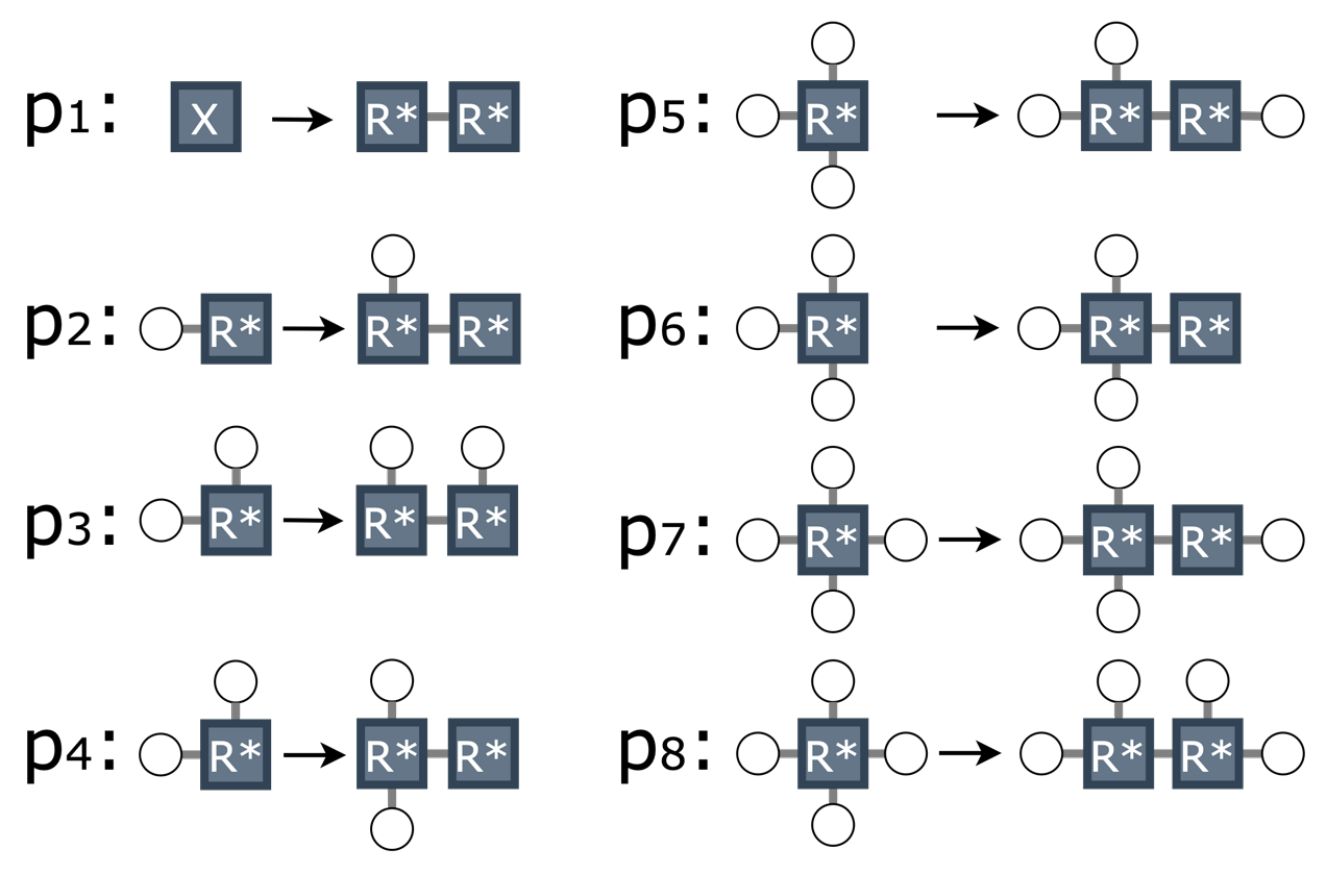}
    \subcaption{$4$-degree meta rules.}
    \label{fig:4deg-grammar}
    \includegraphics[width=0.6\linewidth]{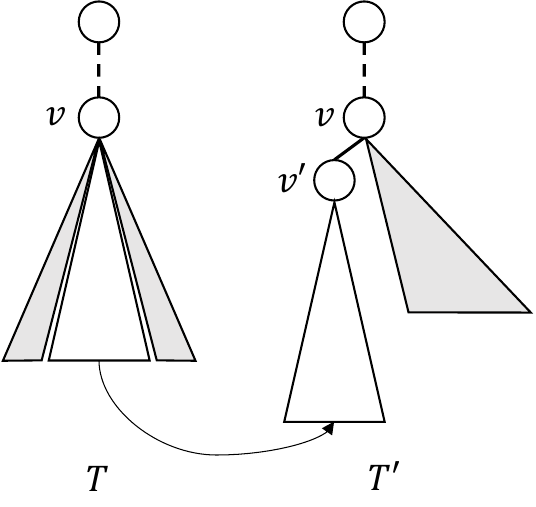}
    \subcaption{One-step inserting operation.}
    \label{fig:insertion}
\end{minipage}

\caption{Illustration of $4$-degree meta grammar and one-step inserting operation.}
\vspace{-4.0ex}
\end{wrapfigure}

\textbf{Meta Rules Construction.}
We refer to the production rule set of $k$-degree, edit-complete, and minimal meta grammar as a \emph{$k$-degree meta rule set}.
We visualize $4$-degree meta rules in Figure~\ref{fig:4deg-grammar}, where $\{p_1\}$, $\{p_1, p_2, p_3\}$, and $\{p_1, p_2, p_3, p_4, p_5\}$ correspond to meta rule sets of degree $1$ to $3$, respectively.
$1$-degree meta rule set $\Pmc_{\metagrammar}^{(1)}$ is constructed as
\begin{align*}
&\Pmc_{\metagrammar}^{(1)} = \Pmc^{(1)} = \{p_0^{(1)}\},\\
& p_0^{(1)}: \mathit{LHS}_{0}^{(1)} \rightarrow \mathit{RHS}_{0}^{(1)},  \mathit{LHS}_{0}^{(1)} := (\{\mathcal{X}\}, \varnothing),\  \mathit{RHS}_{0}^{(1)} :=H(V_{R, 0}^{(1)},E_{R, 0}^{(1)}), \\
& V_{R,0}^{(1)} = \{\Rmcstar_1, \Rmcstar_2\}, E_{R,0}^{(1)} = \{(\Rmcstar_1, \Rmcstar_2)\}.
\end{align*} 
When using the meta grammar for production, we treat $\Rmcstar_1, \Rmcstar_2$ as the same type of non-terminal node, i.e. $\Rmcstar = \Rmcstar_i, i=1, 2$, despite the use of indices in the above formulation.

For $k$-degree meta rule set $\Pmc_{\metagrammar}^{(k)}$ ($k>1$), the construction is achieved by induction:
\begin{align*}
&\Pmc_{\metagrammar}^{(k)} = \Pmc_{\metagrammar}^{(k-1)} \cup \Pmc^{(k)}, \ \Pmc^{(k)}=\{p_0^{(k)}\} \cup \bigcup_{i=1}^{\lfloor \frac{k}{2} \rfloor} \{p_i^{(k)}\}, \\
& p_0^{(k)}: \mathit{LHS}_{0}^{(k)} \rightarrow \mathit{RHS}_{0}^{(k)}, \ \mathit{LHS}_{0}^{(k)}:=H(V_{L,0}^{(k)}, E_{L,0}^{(k)}), \ \mathit{RHS}_{0}^{(k)}:=H(V_{R,0}^{(k)}, E_{R,0}^{(k)}), \\
& p_i^{(k)}: \mathit{LHS}_{i}^{(k)} \rightarrow \mathit{RHS}_{i}^{(k)}, \ \mathit{LHS}_{i}^{(k)}:=H(V_{L,i}^{(k)}, E_{L,i}^{(k)}), \ \mathit{RHS}_{i}^{(k)}:=H(V_{R,i}^{(k)}, E_{R,i}^{(k)}), i = 1,...,\lfloor \frac{k}{2} \rfloor,\\
& V_{L,0}^{(k)} = \{\Rmcstar\} \cup \bigcup_{j=1}^{k-1}\{V_{anc,j}\}, \ E_{L,0}^{(k)} = \bigcup_{j=1}^{k-1}\{(\Rmcstar, V_{anc,j})\}, \\
& V_{R,0}^{(k)} = \{\Rmcstar_1, \Rmcstar_2\} \cup \bigcup_{j=1}^{k-1}\{V_{anc,j}\}, \ E_{R,0}^{(k)} = \{(\Rmcstar_1, \Rmcstar_2)\} \cup \bigcup_{j=1}^{k-1}\{(V_{anc,j}, \Rmcstar_2)\}, \\
& V_{L,i}^{(k)} = \{\Rmcstar\} \cup \bigcup_{j=1}^{k}\{V_{anc,j}\}, \ E_{L,i}^{(k)} = \bigcup_{j=1}^{k}\{(\Rmcstar, V_{anc,j})\},\\
& V_{R,i}^{(k)} = \{\Rmcstar_1, \Rmcstar_2\} \cup \bigcup_{j=1}^{k}\{V_{anc,j}\}, \ E_{R,i}^{(k)} = \{(\Rmcstar_1, \Rmcstar_2)\} \cup \bigcup_{j=1}^{i}\{(V_{anc,j}, \Rmcstar_1)\} \cup \bigcup_{j=i+1}^{k}\{(V_{anc,j}, \Rmcstar_2)\}.
\end{align*}

Specifically, the $k$-degree meta rule set $\Pmc_{\metagrammar}^{(k)}$ contains all the rules from $(k{-}1)$-degree meta rule set $\Pmc_{\metagrammar}^{(k-1)}$ as well as other newly introduced rules $\Pmc^{(k)}$.
Each rule contains one non-terminal node $\Rmcstar$ on the $\mathit{LHS}$ and two non-terminal nodes $\Rmcstar_1$ and $\Rmcstar_2$ on the $\mathit{RHS}$.
The $\mathit{LHS}$ of $p_0^{(k)}$ has a degree of $k{-}1$ as it contains $k{-}1$ anchor nodes $V_{anc}$.
The $\mathit{RHS}$ of $p_0^{(k)}$ attaches one non-terminal node $\Rmcstar_2$ to the other non-terminal node $\Rmcstar_1$ which connects all the anchor nodes.
Therefore, the $\mathit{RHS}$ has a degree of $k$ which is larger than the degree of the $\mathit{LHS}$ by one.
In all the other rules, the $\mathit{LHS}$ has $k$ anchor nodes and is of degree $k$ while the $\mathit{RHS}$ keeps a maximal degree of $k$. 
The $k$ anchor nodes on the $\mathit{RHS}$ are combinatorially distributed to $\Rmcstar_1$ and $\Rmcstar_2$, i.e. if $\Rmcstar_1$ is attached to $i$ anchor nodes, $\Rmcstar_2$ is attached to the rest of $k{-}i$ anchor nodes.
Since we consider $\Rmcstar_1$ and $\Rmcstar_2$ to be the same during grammar production, the range of $i$ is $\{1,...,\lfloor \frac{k}{2} \rfloor\}$.
In total, there are $1{+}\lfloor \frac{k}{2} \rfloor$ rules in $\Pmc^{(k)}$. 

\emph{Proof of Edit Completeness.} 
According to the definition in~\cite{zhang1996constrained, paassen2018revisiting}, tree edit distance $\mathit{TED}(T, T')$ between two trees $T$ and $T'$ is the minimum number of operations required to transform one tree into the other.
There are three types of available operations: inserting, deleting, and relabeling.
Consider two arbitrary trees $T$ and $T'$ that satisfy the definition of edit completeness (being derived from the meta grammar, $\mathit{TED(T, T')} = 1$, and $|T| < |T'|$). 
Since all the trees generated from the meta grammar have homogeneous tree node labels (i.e. $\Rmcstar$), there is no relabeling operation in $\mathit{TED}$ calculation. 
Furthermore, the tree size relation between $T$ and $T'$ admits only one type of tree edit operation: a one-step inserting.
Figure~\ref{fig:insertion} illustrates the one-step inserting operation.
It operates on a certain node $v$ in $T$ whose subtree is $T_{sub}(v)$. 
A new node $v'$, which takes a subset of $T_{sub}(v)$ as the children, is added as a new child of node $v$.
The rest of $T_{sub}(v)$ stays as the children of node $v$.
Suppose the degree of node $v$ in $T$ is $m$ and the degree related to subtree $T_{sub}(v)$ is $n$.
Without loss of generality, we suppose $0\leq n \leq m/2$.
We can treat $v$ in $T$ as the $\Rmcstar$ on the $\mathit{LHS}$, $v'$ and $v$ in $T'$ as $\Rmcstar_1$ and $\Rmcstar_2$ on the $\mathit{RHS}$, respectively.
Then the one-step inserting operation can be achieved using the rule $p_{n}^{\smash{(m+1)}} \in \Pmc^{(m+1)}$ from $(m{+}1)$-degree meta rule set $\Pmc_{\metagrammar}^{(m+1)}$ as constructed above.
Since $\Delta(T')\geq m{+}1$, the meta grammar $\Pmc_{\metagrammar}^{(k)}$ that generates $T'$ has $k\geq m{+}1$. So $p_{n}^{(m+1)} \in \Pmc^{(m+1)} \subseteq \Pmc_{\metagrammar}^{(k)}$; thus finish the proof.

\emph{Proof of Degree $k$.} 
For an arbitrary tree $T$ with maximal degree $k$, we prove that meta grammar $\Pmc_{\metagrammar}^{(k)}$ can generate $T$ by finding a reverse path of production rule from the tree $T$ to the initial node $\mathcal{X}$.
We first find all the nodes $\{v_i\}_{i=1}^{n}$ in $T$ that have a degree of $k$.
For each node, by edit completeness, we can use the inverse operation of inserting and make the node to have a degree of $k{-}1$.
It can be shown from the proof for edit completeness that each inverse operation of inserting corresponds to an inverse deployment of a production rule.
By using $n$ reverse production steps on all the $k$-degree nodes, we can obtain a $T'$ with $\Delta(T')=k{-}1$.
We can continue the process until the tree is $0$-degree, i.e. the $\mathcal{X}$; thus finish the proof.

\emph{Proof of Minimality.} 
From edit completeness, we know each production rule in the meta grammar corresponds to at least one case of one-step inserting operation.
As constrainted by degree $k$, arbitratry trees that are of degree smaller than $k$ need to be covered by the meta grammar.
We thus need to handle all the possible cases of one-step inserting operation when finding the reverse production path from the tree to $\mathcal{X}$.
Note that one-step inserting opeartion involves the division of a subtree, which has combinatorial ways.
Then every combinatorial distribution of anchor nodes corresponds to one case of subtree division. 
This proves each production rule in the meta rule set is necessary.
Since there are no duplicate rules, it can be concluded that the meta grammar constructed above is minimal.

\textbf{$\mathbf{4}$-degree Meta Rules.} 
Figure~\ref{fig:4deg-grammar} shows the ${4}$-degree meta grammar we use in our approach.
Since the edit completeness is the foundation to prove all the other attributes of the meta grammar, we use a tree edit distance toolkit\footnote{\url{https://pythonhosted.org/zss/}} and verify its correctness for the ${4}$-degree meta grammar.

\section{Molecular Rules Construction}\label{app:molrules}

\begin{algorithm}[h!]
\LinesNumbered
\caption{Molecular Rules Construction.}
\label{alg:molerule}
\newcommand\mycommfont[1]{\footnotesize\ttfamily\textcolor{blue}{#1}}
\SetCommentSty{mycommfont}
\SetKwInOut{Input}{Input}
\SetKwInOut{Output}{Output}
\SetKwFunction{FCC}{ConnectedComponents}
\SetKwFunction{FGN}{GetNeighbors}
\SetKwFunction{FInRing}{InRing}
\SetKwFunction{FRule}{RuleConstruction}
\Input{molecular hypergraph $H=(V,E_H)$, probability function $\phi(\cdot;\theta)$}
\Output{molecular rule set $\mathcal{P}_{mol}=\{p_i\}_{i=1}^{M}$, junction tree $T=(V_T, E_T)$ where each $v_T\in V_T$ is a subset of $V$}
Initialize $\mathcal{P}_{mol}=\varnothing$, $V_T=\varnothing$, $E_T=\varnothing$;\\
Create a set for unvisited nodes $V_{unv}\gets V$, a set for unvisited hyperedges $E_{unv}\gets E_H$;\\
\While{$E_{unv} \neq \varnothing$}{
$\mathbf{e} = \begin{bmatrix}e_1,...,e_K\end{bmatrix}, e_i\in E_{unv}, i=1,...,K{=}|E_{unv}|$;\\
$\mathbf{X}\sim\text{Bernoulli}(\phi(\mathbf{e};\theta))$;\\
\tcc{Construct a rule for each connected component}
\For{{\normalfont{each}} $H_{sub}=(V_{sub}, E_{sub})$ {\normalfont{{in}}} \FCC{$\mathbf{X}$, $H$}}{
    \tcc{Separate the connected component}
    $\hat{V}_{sub} \gets V_{sub}$, $\hat{E}_{sub} \gets E_{sub}$, $V_{anc}\gets \varnothing$, $E_{anc} \gets \varnothing$;\\
    \For{{\normalfont{each}} $v\in V_{sub}$}{
        $V_{n}=$\ \FGN{$v$, $H$};\\
        \uIf{$V_{n}\cap V_{sub} \neq \varnothing$}{
            \uIf{$v\notin V_{unv}$ {\normalfont{or}} \FInRing{$v$, $H$}}{
                $\hat{V}_{sub} \gets \hat{V}_{sub} \setminus \{v\}$;\\
                $\hat{E}_{sub} \gets \hat{E}_{sub} \setminus \{(s,v)|(s,v)\in E_{sub}\}$; \\
                $V_{anc} \gets V_{anc} \cup \{v^{(i)}_{anc}, i=|V_{anc}|+1\}$;\\
                $E_{anc} \gets E_{anc}\cup\{(s,v^{(i)}_{anc})|(s,v)\in E_{sub}, i=|V_{anc}|+1\}$;\\
            }\uElse{
                $V_{anc} \gets V_{anc} \cup \{v^{(i)}_{anc}, i=|V_{anc}|+k\}_{k=1}^{|V_{n}\cap V_{sub}|}$;\\
                $E_{anc} \gets E_{anc}\cup\{(s,v^{(i)}_{anc})|(s,v)\in E_{sub}, i=|V_{anc}|+k\}_{k=1}^{|V_{n}\cap V_{sub}|}$;\\
            }
        }
    }
    \tcc{Construct a production rule}
    $p=$ \FRule{$\hat{V}_{sub}$, $\hat{E}_{sub}$, $V_{anc}$, $E_{anc}$};\\
    $\mathcal{P}_{mol}\gets\mathcal{P}_{mol}\cup\{p\}$;\\
    \tcc{Construct junction tree}
    $v_T = (V_{sub})$;\\
    \For{{\normalfont{each}} $ \{v_t\in V_T| v_t\cap v_T \neq \varnothing\}$}{
        $E_T \gets E_T \cup \{(v_t, v_T)\}$;
    }
    $V_T\gets V_T\cup\{v_T\}$;\\
    \tcc{update visited status of nodes and hyperedges}
    $V_{unv}\gets V_{unv}\cup V_{sub}$, $E_{unv}\gets E_{unv}\cup E_{sub}$;
}
}
\KwRet$\mathcal{P}_{mol}$, $T=(V_T, E_T)$;
\end{algorithm}

To obtain the input for the molecular rule construction, we first convert the molecule into a molecular hypergraph $H=(V,E_H)$.
A node $v\in V$ represents an atom of the molecule.
A hyperedge $e\in E_H$ corresponds to either a bond that joins only two nodes or a ring (including aromatic ones) that joins all nodes in the ring.
An illustration is provided in Figure 2 of~\cite{guo2022data}.
The probability function $\phi(\cdot;\theta)$ is defined on each hyperedge: $\phi(e;\theta) = \sigmoid(-\mathcal{F}_{\theta}(f(e)))$, where $\sigmoid(\cdot)$ is the sigmoid function, $\mathcal{F}_{\theta}(\cdot)$ is a two-layer fully connected network whose final output dimension is $1$, and $f(\cdot)$ is a feature extractor using a pre-trained GNN~\citep{hu2019strategies}.

Algorithm~\ref{alg:molerule} illustrates the detailed process of constructing molecular rules for a single molecule.
Note that in our approach, the molecular rule construction is performed simultaneously for all input molecules.
For each molecule,
we first perform an i.i.d. sampling on all the hyperedges following a Bernoulli distribution which takes the value $1$ with probability indicated by $\phi(e;\theta)$.
We then obtain a binary vector $\mathbf{X}$ that indicates whether each hyperedge is sampled (line $4$-$5$).
Next, all connected components are extracted with respect to the sampled hyperedges (line $6$).
A production rule is constructed for each connected component $H_{sub}=(V_{sub}, E_{sub})$ (line $7$-$19$).
Specifically, a production rule needs two components: a hypergraph $\hat{H}_{sub} = (\hat{V}_{sub}, \hat{E}_{sub})$ for $\mathit{RHS}$ and anchor nodes $V_{anc}$ indicating the correspondence between the $\mathit{LHS}$ and the $\mathit{RHS}$.
$\hat{H}_{sub}$ is obtained by removing visited nodes and in-ring nodes from $H_{sub}$.
$V_{anc}$ contains nodes from $V$ that are connected to nodes from $H_{sub}$ but do not appear in $H_{sub}$ themselves.
$E_{anc}$ are the edges connecting anchor nodes to $\hat{H}_{sub}$ following the same connectivity in the original graph $H$.
Following~\cite{guo2022data}, the function {\footnotesize\ttfamily{RuleConstruction(}}$\hat{V}_{sub}$, $\hat{E}_{sub}$, $V_{anc}$, $E_{anc}${\footnotesize\ttfamily{)}} returns a production rule $p: \mathit{LHS} \rightarrow \mathit{RHS}$ constructed as 
\begin{equation}
\begin{aligned}
\mathit{LHS} := \ H(V_L,E_L), \ &V_L=\{\mathcal{R}^*\}\cup V_{anc}\ , \ E_L=\{(\mathcal{R}^*, v) |v\in V_{anc}\}\ ,\\
\mathit{RHS} := \ H(V_R,E_R), \ &V_R=\hat{V}_{sub}\cup V_{anc}\ , \ E_R= \hat{E}_{sub}\cup E_{anc}.
\end{aligned}
\end{equation}
In contrast to \cite{guo2022data}, our molecular rule construction does not replace the connected component with the non-terminal node $\mathcal{R}^*$ at every iteration since according to our definition, the molecular rule does not contain any non-terminal nodes on the $\mathit{RHS}$.

The junction tree of the molecule is constructed along with the construction of molecular rules (line $21$-$24$). 
We simply treat each connected component $V_{sub}$ sampled at each iteration as a node $v_T \in V_T$ in the junction tree $T=(V_T, E_T)$.
Two nodes are connected by an edge $e_T\in E_T$ if their corresponding connected components share hypergraph nodes from $H$.
Since each hyperedge in $H$ is only visited once, the constructed junction tree then contains all the hyperedges and nodes of $H$ without redundancy \citep{kajino2019molecular}.


\section{Grammar-induced Geometry Construction}\label{app:grammardag}
\begin{algorithm}[h]
\LinesNumbered
\caption{Grammar-induced Geometry Construction.}
\label{alg:grammardag}
\newcommand\mycommfont[1]{\footnotesize\ttfamily\textcolor{blue}{#1}}
\SetCommentSty{mycommfont}
\SetKwInOut{Input}{Input}
\SetKwInOut{Output}{Output}
\SetKwFunction{FQ}{.equeue}
\SetKwFunction{FDQ}{.dequeue}
\SetKwFunction{FIso}{GetIsomorphicGraph}
\SetKwFunction{FIsIso}{IsIsomorphic}
\Input{meta production rules $\Pmc_{\metagrammar}=\{p_i\}_{i=1}^{N}$, maximum BFS depth $D$, a set of molecular hypergraphs $\{H_i\}_{i=1}^{M}$ and their corresponding junction trees $\mathcal{J}=\{T_i\}_{i=1}^{M}$}
\Output{geometry in the form of a graph $\mathcal{G}=(\mathcal{V}, \mathcal{E})$ where each $v=H_v=(V_v, E_v) \in\mathcal{V}$ represents a meta tree or a molecular hypergraph and $\mathcal{E}$ is the edge set of the geometry}
\tcc{Add root of the geometry }
Initialize $v_{root} = H_{root} = (\mathcal{X}, \varnothing)$, $\mathcal{V} = \{v_{root}\}$, $\mathcal{E} = \varnothing$;\\
Create a queue data structure $Q$;\\
$Q$\FQ{$v_{root}$};\\
\tcc{Breadth-first search for meta geometry, pre-computed offline}
\While{$Q$ {\normalfont{is not empty}} {\normalfont{and}} $|\mathcal{V}| \leq D$}{
    $v = Q$\FDQ{};\\
    \tcc{Expand the meta geometry}
    \For{{\normalfont{each}} $p_i\in \Pmc_{\metagrammar}$}{
        \uIf{$p_i$ {\normalfont{is applicable to}} $H_v$}{
            $H_v{\pito} H_{new}$;\\ 
            $v_{new} =  H_{new}$;\\
            \uIf{$v_{new} \notin \mathcal{V}$}{
                $\mathcal{V} \gets \mathcal{V} \cup \{v_{new}\}$, $\mathcal{E} \gets \mathcal{E} \cup \{(v, v_{new})\}$;\\
                $Q$\FQ{$v_{new}$};\\
            }
            \uElse{
                $\hat{v}=$\ \FIso{$H_{new}$, $\mathcal{V}$},
                $\mathcal{E} \gets \mathcal{E} \cup \{(v, \hat{v})\}$;\\
            }
        }
    }
}

\tcc{Connect molecular leaves during run-time}
\For{{\normalfont{each}} $T_i \in \mathcal{J}$}{
    \For{{\normalfont{each}} $v \in \mathcal{V}$}{
        \uIf{\FIsIso{$H_v$, $T_i$}}{
            $v_i = H_i$;\\
            $\mathcal{V} \gets \mathcal{V} \cup \{v_i\}$, $\mathcal{E} \gets \mathcal{E} \cup \{(v, v_i)\}$;\\
        }
    }
}
\KwRet $\mathcal{G}=(\mathcal{V}, \mathcal{E})$;
\end{algorithm}
Algorithm~\ref{alg:grammardag} illustrates the detailed algorithm to construct the grammar-induced geometry.
It contains two parts: the construction of the meta geometry (line $4$-$14$) and the construction of the molecular leaves (line $15$-$19$).
The meta geometry construction follows the standard breadth-first search (BFS) starting from the root $H_{root} = (\mathcal{X}, \varnothing)$ (line $1$).
Every time when we visit a node $v\in\mathcal{V}$ in $\mathcal{G}$, we find all the rules that are applicable to $H_v$ from the meta rule set (line $6$-$7$).
Each applicable rule is applied to $H_v$ in order to create a new meta tree $H_{new}$ (line $8$).
Depending on whether there is a node in $\mathcal{V}$ that represents a isomorphic tree to $H_{new}$ in the current geometry, we either create a new node $v_{new}$ (line $9$) or find the existing matched node $\hat{v}$ (line $14$).
We then add an edge between $v$ and $v_{new}$, or between $v$ and $\hat{v}$ (line $11$ and $14$).
The function {\footnotesize\ttfamily{GetIsomorphicGraph(}}$H_{new}$, $\mathcal{V}${\footnotesize\ttfamily{)}} enumerates every node in $\mathcal{V}$, checks if the tree represented by the node is isomorphic to $H_{new}$, and returns the matched node.
We use the algorithm from \cite{cordella2001improved} which is implemented in {\footnotesize\ttfamily{networkx}}\footnote{\url{https://networkx.org/documentation/stable/reference/algorithms/isomorphism.html}} for graph isomorphism test.
To increase the speed of {\footnotesize\ttfamily{GetIsomorphicGraph(}}$H_{new}$, $\mathcal{V}${\footnotesize\ttfamily{)}}, we use Weisfeiler Lehman graph hash~\citep{shervashidze2011weisfeiler} and only perform isomorphism test for graph pairs that share the same hashing code.

The molecular leaves are added during run-time (line $15$-$19$).
For each input molecular hypergraph, we check if there is a meta tree node that is isomorphic to its junction tree.
If so, we add an edge connecting the matched meta tree node to the molecular hypergraph.
The function {\footnotesize\ttfamily{IsIsomorphic(}}$H_v$, $T_i${\footnotesize\ttfamily{)}} is implemented using the same package for {\footnotesize\ttfamily{GetIsomorphicGraph(}}$H_{new}$, $\mathcal{V}${\footnotesize\ttfamily{)}} and returns true if $H_v$ and $T_i$ are isomorphic to each other.


\section{Optimization}\label{app:optimization}

\begin{wrapfigure}{r}{0.5\textwidth}
    \centering
    \begin{tikzpicture}[node distance={18mm}, main/.style = {draw, circle}] 
    \node[main] (1) {$\mathbf{X}$}; 
    \node[main] (2) [below left of=1]{$\theta$}; 
    \node[main] (3) [right of=1]{$\mathbf{u}_T$}; 
    \node[main] (4) [below left of=3]{$\varphi$};
    \node[main] (5) [below of=3]{$\psi$};
    \node[main] (6) [below right of=3]{$\alpha$};
    \node[main] (7) [right of=3]{${l}(\cdot, \cdot)$};
    \node[main] (8) [above of=7]{$\hat{\mathbf{u}}$};
    \draw[->] (2) -- (1);
    \draw[->] (1) -- (3);
    \draw[->] (4) -- (3);
    \draw[->] (5) -- (3);
    \draw[->] (6) -- (3);
    \draw[->] (3) -- (7);
    \draw[->] (8) -- (7);
    \end{tikzpicture} 
    \caption{A graphical model of dependency.}
    \label{fig:param}
\end{wrapfigure}
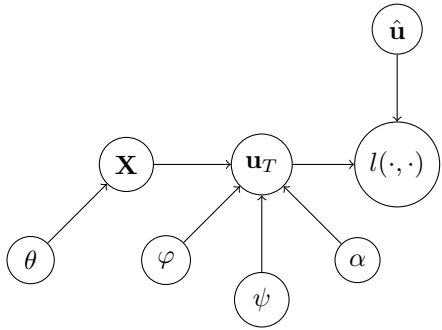

Figure \ref{fig:param} shows a graphical model of dependency in our optimization problem.
The two sets of parameters $\theta$ and $(\varphi, \psi, \alpha)$ are independent and thus can be iteratively optimized using block coordinate descent.
Since the objective in Equation~\ref{eq:opt} is differentiable with respect to $(\varphi, \psi, \alpha)$, we can use gradient descent for the optimization iterations related to the graph neural diffusion.
The geometry learning, however, is non-differentiable due to the fact that we perform sampling to construct molecular rules and construct the geometry.
Hence, we rewrite the objective for optimizing $\theta$ in an expectation form and apply REINFORCE ~\citep{williams1992simple} to obtain a stochastic gradient, as done in~\cite{guo2022data}:
\begin{align*}
\min_{\theta}{l}(\mathbf{u}_T,  \hat{\mathbf{u}}) &= \min_{\theta}\mathbb{E}_{\mathbf{X}}\Big[{l}(\mathbf{u}_T,  \hat{\mathbf{u}})\Big],\\
\nabla_{\theta} \mathbb{E}_{\mathbf{X}}\Big[{l}(\mathbf{u}_T,  \hat{\mathbf{u}})\Big]
&= \int_{\mathbf{X}} {l}(\mathbf{u}_T,  \hat{\mathbf{u}})\nabla_{\theta} p(\mathbf{X}) \\
&= \mathbb{E}_{\mathbf{X}}\Big[ {l}(\mathbf{u}_T,  \hat{\mathbf{u}})\nabla_{\theta}\log(p(\mathbf{X}))\Big]
\approx \frac{1}{N}\sum_{n=1}^{N} {l}(\mathbf{u}^{(n)}_T,  \hat{\mathbf{u}}) \nabla_{\theta}\log(p(\mathbf{X}^{(n)})),
\label{eq:reinforce}
\end{align*}
where $\mathbf{X}$ is a concatenation of binary vectors indicating how hyperedges are sampled in molecular rule construction in Appendix~\ref{app:molrules}.

\section{Examples of Retro-Synthesis Paths}\label{app:retro}
\begin{figure*}[t] 
\centerline{\includegraphics[width=1.0\linewidth]{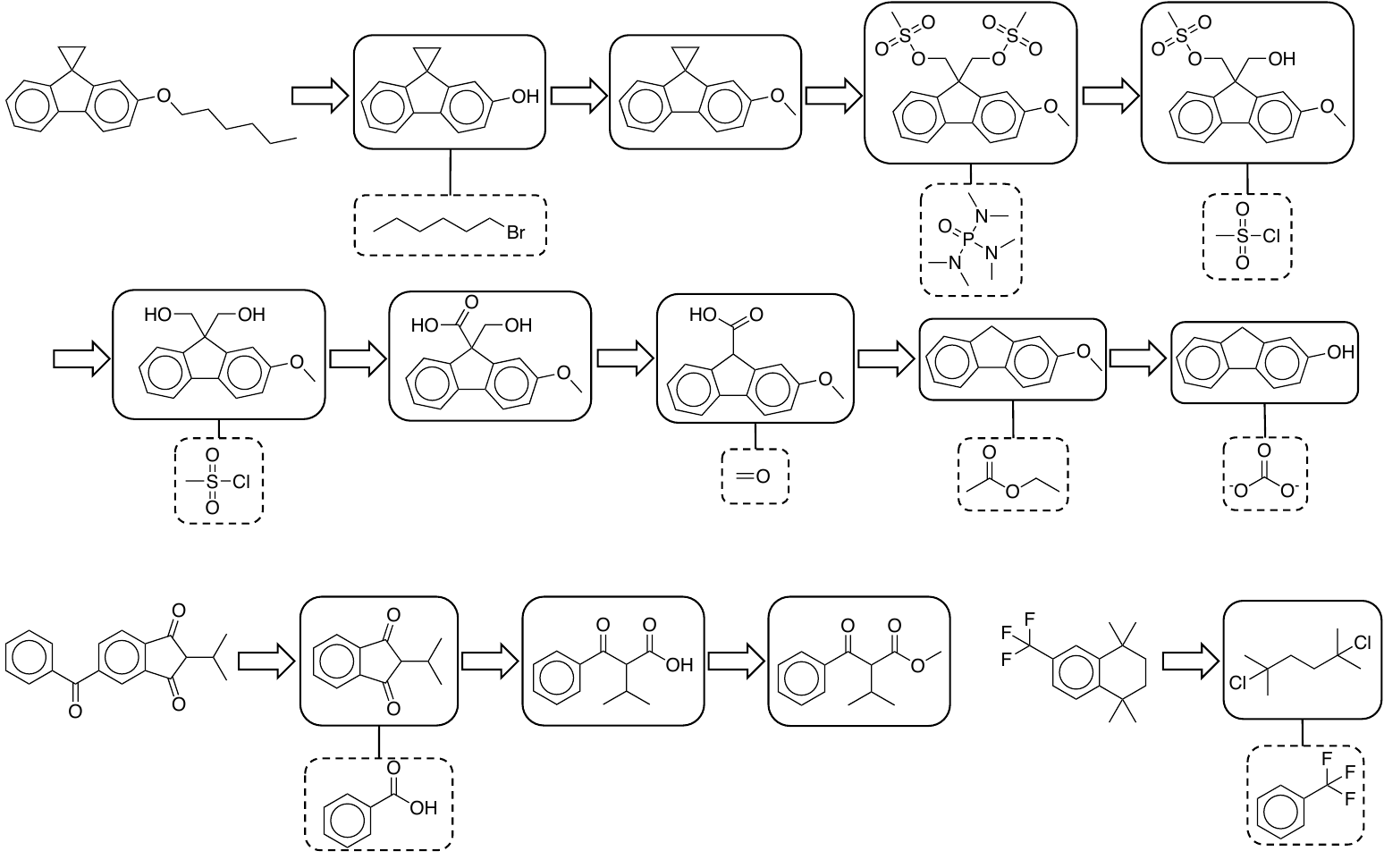}}
\vspace{-1.5ex}
\caption{
Examples of retro-synthesis paths. 
}
\label{fig:retro}
\vspace{-2.5ex}
\end{figure*}

Figure~\ref{fig:retro} shows the retro-synthesis paths of three molecules generated using our pipeline. 
Since Retro$^*$ score is used as one of the grammar metrics, our approach is capable of retro-synthesis planning for all the generated molecules, providing a complete pipeline for novel molecule discovery.

\end{document}